\documentclass[3p,authoryear]{article}
\usepackage{amssymb}
\usepackage{natbib}
\usepackage[usenames, dvipsnames]{color} 
\usepackage[margin=1in]{geometry}

\usepackage{arydshln}
\usepackage[normalem]{ulem}
\usepackage{amsmath}
\usepackage{lscape}
\usepackage{algorithmicx}
\usepackage{algpseudocode}
\usepackage{url}
\usepackage{graphicx,xcolor}

\usepackage{subcaption}
\usepackage[linesnumbered,ruled,vlined]{algorithm2e}
\usepackage{booktabs}
\usepackage{tikz}
\usetikzlibrary{positioning, shapes, shadows, arrows}
\usetikzlibrary{decorations.pathreplacing}
\tikzstyle{abstract}=[circle, draw=black, fill=white]
\tikzstyle{labelnode}=[circle, draw=white,opacity=.2,text opacity=1]
\tikzstyle{invisiblenode}=[circle,dashed, inner sep=1pt,circle split,line width=1mm,minimum size=1.5cm]
\tikzstyle{line} = [draw, -latex']

\definecolor{red}{gray}{0.85}
\definecolor{green}{gray}{0.8}
\definecolor{blue}{gray}{0.7}
\definecolor{yellow}{gray}{0.6}
\definecolor{violet}{gray}{0.5}
\definecolor{orange}{gray}{0.4}
\definecolor{brown}{gray}{0.3}

\title{A simple and efficient architecture for trainable activation functions}

\author{Andrea Apicella, Francesco Isgr\`o and Roberto Prevete}
\date{Dipartimento di Ingegneria Elettrica e delle Tecnologie dell'Informazione\\Universit\`a di Napoli Federico II}
\begin{document}
\maketitle

%
\begin{abstract}
Learning automatically the best activation function for the task is an active topic in neural network research.
At the moment, despite promising results, it is still difficult to determine a method for learning an activation function that is at the same time theoretically simple and easy to implement. 
Moreover, most of the methods proposed so far introduce new parameters or \emph{adopt} different learning techniques. 
In this work we propose a simple method to obtain trained activation function which adds to the neural network local subnetworks with a small amount of neurons.
Experiments show that this approach could  lead to better result with respect to using a pre-defined activation function, without introducing a large amount of extra parameters that need to be learned. 
\end{abstract}

\textbf{keywords:}
neural networks, machine learning, activation functions, adaptive activation functions




\section{Introduction}
\label{sec:introduction}
The success of deep learning approaches has led to an increase in interest in MultiLayer FeedForward (MLFF) neural networks. 
MLFF networks are composed of $N$ elementary computing units (neurons), which are organized in $L>1$ layers.
The first layer of a MLFF network is composed of $d$ input variables. 
Each neuron $i$ belonging to a layer $l$, with $l>1$, \textbf{receives possibly connections} from all the neurons (or input variables in case of $l=2$) of the previous layer $l-1$. 
Eeach connection is associated to a real value called \textit{weight}. The flow of computation proceeds from the first layer to the last layer (\textit{forward propagation}).  
The last neuron layer is called \textit{output} layer, the remaining neuron layers are called \textit{hidden} layers. 
The computation of a neuron $i$ belonging to the layer $l$ corresponds to a two-step process: first is computed the neuron activation $a_{i}^l$ and then the neuron output $z_i^{l}$. The neuron activation $a_{i}^l$ is usually constructed as a linear combination of the outputs of the previous layer: $a_{i}^{l}=\sum_j w_{ij}^l z^{l-1}_j + b^l_{i}$ where $w_{ij}^l$ is the weight of the connection going to the neuron $j$ belonging to the layer $l-1$ to the neuron $i$, $b^l_{i}$ is a parameter said $bias$, $l=1,..,L$ and $j$ runs on the indexes of the neurons of the layer $l-1$ which send connections to the neuron $i$. If $l=2$ the the variables $z^{l-1}$ correspond to the input variables. The neuron output $z_{i}^l$ is usually computed by a differentiable, non linear activation function $f(\cdot)$: $z_{i}^l=f(a_{i}^l)$. The nonlinear functions $f(\cdot)$ are generally chosen as simple \textit{fixed} functions such as the \textit{logistic sigmoid} or the \textit{tanh} function. 

Given a MLFF network with $d$ input variables and $c$ neurons in the output layer, it achieves a functional mapping from a $d$-dimensional space to a $c$-dimensional space. Thus, a MLFF network can be interpreted as a non-linear parametric function $\textbf{y}=Net(\textbf{x};\boldsymbol{\theta})$, where the parameters $\boldsymbol{\theta}$ are all the weigths and biases of the network and $\textbf{y}$ is the response of the output layer.  
The approximation properties of MLFF networks have been widely studied \citep{devore1996}.
In a nutshel, a function approximation problem can be summarized as follows \citep{bishop2006,ripley2007}:
given an unknown function $F: \textbf{x} \in R^d \longrightarrow \textbf{y}=F(\textbf{x}) \in R^c$ and a data set  $\{(\textbf{x}^n,\textbf{t}^n)\}_{n=1}^N$ representing a sampling of the unknown function, where $\textbf{t}^n= F(\textbf{x}^n)+\boldsymbol{\epsilon}$, usually called \textit{targets}, are the values assumed by $F$ in $\textbf{x}^n$ with added an unknown noise $\boldsymbol{\epsilon}$, the task is to find the appropriate values of the parameters of a parametric function $M$ so as to get as close as possible to the unknown function $F$. In this context, there are two different problems, the first one concerns the expressive power of the parameterized function $M$, that is, if there are parameter values for which it is possible to approximate the unknown function $F$, the second one concerns the possibility of actually finding such parameter values. 
Interestingly, regarding the first problem, a MLFF network with a single hidden layer, which is usually called \textit{shallow} network, can approximate arbitrarily well any functional continuous mapping defined on a compact input domain, provided the number of hidden neurons is sufficiently large and the activation functions of the hidden neurons satisfy suitable properties, for example, to be  sigmoidal or, more in general, not-polynomial functions \citep{sonoda2017,bishop2006,pinkus1999}.
In other words, given a certain  desired degree of approximation, it exists a set of parameters $\bar{\boldsymbol{\theta}}$ for which the neural network $Net(\textbf{x};\bar{\boldsymbol{\theta})}$ approaches the unknown function within this degree of approximation, supposed to have a sufficient number of hidden neurons and appropriate activation functions. In this sense, 
MLFF networks are said to be \textit{universal approximators}. 

However, the \textit{key problem} is how to find these suitable network parameters, i.e., weights and biases.  The process to determine the values of weights and biases on the basis of the data set is called \textit{learning} or \textit{training}. Importantly, although the choice of the non linear activation functions $f(\cdot)$ does not affect the MLFF network's universal approximator property, provided certain constraints are satisfied, this choice becomes a key aspect when network weights and biases are to be found during the training process. To clarify this aspect, let us briefly summarize what is a training process. The training process generally corresponds to the minimization of an \textit{error function} with respect to the network parameters. The error function typically assumes the following form (although many other forms are possible \citep{bishop2006} ):  $E(\boldsymbol{\theta})=\cfrac{1}{2}\sum_{n=1}^{N} \sum_{k=1}^c [y_k(\textbf{x}^n;\boldsymbol{\theta})-t^n_k]^2$ where $y_k(\textbf{x}^n;\boldsymbol{\theta})$ represents the output of the neuron $k$ belonging to the output layer as a function of both the input $\textbf{x}^n$ and the network parameters $\boldsymbol{\theta}$. The quantity $t^n_k$ represents the target value for output neuron $k$ when the input is $\textbf{x}^n$.The solution for the network parameter values at the \textit{global minimum} of the error function is usually found by iterating a gradient-based algorithm with the gradient computed through backpropagation \citep{bishop2006}. Since for MLFF networks the error function typically will be a highly non-linear function of the parameters (not-convex surface),  there may exist many \textit{local minima} or \textit{saddlepoints}. Notice that parameter regions where the error function is very ``flat'' can mimick local minima insofar as the learning process is "trapped" for very long periods of time. In a learning process the main difficult is to avoid these stationary points or regions of the error function. Thus, the choice of the activation functions has a relevant impact on the shape of the error function and, consequently, on the performance of the learning process. Moreover, this choice can affect the number of hidden neurons and layers necessary to reach the desired degree of approximation \citep{guliyev2016single,eldan2016power}. 

For these reasons, recently, there is a very large literature proposing activation functions that differ from those standards such as sigmoid and tanh. In particular, the introduction of activation functions as ReLU \citep{nair2010} and similar functions, such as Leaky ReLU  and parametric ReLU described respectively in \citep{maas2013} and \citep{he2015}, has contributed to renew the interest of the scientific community for MLFF networks. The use of these new activation functions has been shown to improve significantly the networks in terms of performance and training speed, thanks to properties as no saturation, e.g. \citep{glorot2010}. 
Another great improvement was given in \citep{clevert2015}, where the learning is speeded up introducing the ELU activation function, and more recently in \citep{klambauer2017} with the introduction of SELU units. 

Thus, finding alternative functions that can potentially improve further the results is still an \textit{open field of research}.
Consistent with this perspective, a number of recent papers compare neural architectures with different activation functions, as in \citep{pedamonti2018}, or propose to search appropriate activation functions within a finite set of potentially interesting activation functions, as in \citep{ramachandran2017}.
However, a very recent field of research focuses on the possibility to learn appropriate activation functions from data, in this way one has \textit{adaptable} (or \textit{trainable}) activation functions which are adjusted during the learning phase towards specific functions, allowing the network to exploit the data better (see, for example, \citep{qian2018}). 
Furthermore, any layer of the network could potentially have their own best activation function, increasing the number of choices to make at the design stage. 
On the other side, it is not guaranteed that fixing the same function for each layer is the best choice. 
Thus, a way to tackle the problem is to learn the activation functions from data, together with the other parameters of the network; the idea is to find the \textit{good} activation functions that, together with the other network parameters, provides a \textit{good} model for the data.

In this paper we introduce a new method for learning activation functions in the context of full-connected and convolutional MLFF neural networks. The impact of this method on the performance of the network are experimentally assessed.  
The idea is built upon the possibility to obtain adaptable activation functions in terms of sub-networks with just one hidden layer. 
In a nutshell, each neuron with a non-linear activation function $f$ can be \textit{substituted} with a neuron with an \textit{Identity} activation function which sends its output to a one-hidden layer sub-network with just one output neuron. 
This substitution enables us to obtain ``any'' activation function $f$, since an one-hidden layer neural network can approximate arbitrarily well any functional continuous mapping from one finite-dimensional space to another, provided the number of hidden neurons is sufficiently large and the activation functions of the hidden neurons satisfy suitable properties, for example, to be  sigmoidal or, more in general, not-polynomial functions \citep{sonoda2017,bishop2006,pinkus1999}.
Thus, our neural network architecture with variable activation functions is again a MLFF neural network. Importantly, this property means that any classical approach applicable to MLFF networks can also be directly applied to our architecture with trainable activation functions. Notably, as we will discuss in Section \ref{sec:relatedWork} and \ref{sec:systemArch} our architecture represents a general framework in which several approaches recently proposed in literature can be included. 


The paper is structured as follows.
In the next section we critically discuss our approach with respect to the current literature. 
In Section \ref{sec:systemArch} we introduce our architecture. 
Section \ref{sec:expFramework} is dedicated to the experimental assessment. 
Finally, Section \ref{sec:conclusion} is left to the conclusions.

\section{Related work}
\label{sec:relatedWork}
Over the last years, ReLU has become the standard activation function for deep neural models, surpassing classic functions as sigmoid and tanh used in the past literature thanks to useful properties, such  as the ability to avoid saturation issues\citep{nair2010,glorot2011}. 
Different variations of the ReLU have been proposed \citep{maas2013,konda2014,dugas2000}. 
All these functions are somehow different from ReLU, but once chosen they remain fixed, with the choice of which one to use taken during the design stage, typically in a heuristic way. A partial attempt to overcome this drawback moves in the direction of searching the best activation function from a predefined set \citep{liu1996,yao1999,ramachandran2017}. These techniques are limited by the fact that the functions are not learned, but just selected from a collection of standard functions. Thus, approaches by trainable activation functions
propose more general frameworks. In this direction one can isolate three basic types of approaches: \textit{parameterized standard activation function}, \textit{linear combination of one-variable functions} and \textit{ensemble of standard activation functions}. In Subsection \ref{subsec:parameterizedSAF}, \ref{subsec:linearCombinationBF} and \ref{subsec:ensemble} we will discuss these three types of approaches. In Subsection \ref{subsec:otherApproaches} we will present other types of solutions. Our discussion will be mainly based on three dimensions: 1) how many new parameters are added to the network model, 2) the possibility or not to use standard techniques, within neural network context, for learning the new parameters, such as backpropagation for computing the error function gradient  or sparse methods, 3) the expressive power of the trainable activation functions.  


\subsection{Parameterized standard activation functions}
\label{subsec:parameterizedSAF}
With the expression parameterized standard activation functions we refer to all the functions with a shape that is very similar to a given standard activation function, but whose diversity from the latter comes from a set of trainable parameters. The addition of these parameters therefore requires changes, even minimal ones, in the learning algorithm, for example, in the case of using gradient-based methods with backpropagation, the partial derivatives of the error function respect to these new parameters are needed. A first attempt to have a parameterized activation function is given in \citep{hu1992} where the proposed activation function uses two trainable parameters $\alpha,\beta$ to rule the function shape of a classic sigmoidal function. Similar works on sigmoidal and hyperbolic tangent functions are discussed in  \citep{yamada1992,yamada1992_2,chen1996,singh2003,chandra2004}.
More recently, the authors in \citep{he2015} introduce PReLU, a parametric version of ReLU, which modifies the function shape when the argument is negative. However, the resulting function remains basically a modified version of the ReLU function that can change its shape in a restricted domain. In \citep{clevert2015} ELU function is proposed,  which outperforms the results obtained by ReLU on CIFAR100 dataset, becoming one of the best activation function currently known. However, it needs an external parameter  $\alpha$ to be set. In \citep{trottier2017}  PELU unit is proposed,  where the need to manually set the $\alpha$ parameter is eliminated using two additional trainable parameters. 

In all the approaches previously described, although the number of added parameters for each node is low, the expressive power of the trainable activation functions is limited.  

\subsection{Linear combination of one-variable functions}
\label{subsec:linearCombinationBF}
In this case, activation functions are modelled in terms of linear combinations of one-variable functions. These one-variable functions can in turn have additional parameters.
For example, in \citep{agostinelli2014} each activation function is represented as a linear combination of $S$ hinge-shaped functions. Each hinge-shaped function has just one parameter which regulates the location of the hinge. The number of additional parameters that must be learned when using this approach is $2SM$, where $M$ is the total number of hidden units in the neural network. During the learning phase, the network can be trained using standard methods based on backpropagation.  Any continuous piecewise-linear function can be approximated arbitrarily well provided the number $S$ of hinge-shaped functions is sufficiently large. 

A similar approach has been recently proposed by \citep{scardapane2018}. In this case, the activation function is modelled as a linear combination of $S$ fixed functions, where the $S$ fixed functions are defined in terms of parametric kernel functions. The parameters of the kernel functions are computed before the network learning phase by some heuristic procedure applied on the data set. During the network learning phase the number of additional parameters is $SM$, however for the kernel functions a number of $KS$ parameters must be computed in a prior stage (where $k$ is the number of parameters of the kernel functions).   
In case of a correct choice of the parameters of the kernel functions, any continuous one-to-one function defined on a compact set can be approximated arbitrarily well, 
provided the number of kernel functions is sufficiently large.

In \citep{omer2018}, in the context of random weight artificial neural networks,
a trainable activation function is proposed in terms of a polynomial function of degree $p$. The coefficients of the polynomial function are computed by linear regression. The number of added parameters corresponds to the number $p+1$ of coefficients for each neuron.

\subsection{Ensemble of standard activation functions}
\label{subsec:ensemble}
In this type of approaches, activation functions are defined as an ensemble of a predetermined number of standard activation functions. For example, the authors of \citep{jin2016} designed an activation $S$-shape function composed by three linear functions taking inspiration by Webner-Fechner \citep{fechner1966} and Stevens law \citep{stevens1957}, or in \citep{qian2018} a mixture of eLU and ReLU is presented.
Interestingly, in \citep{sutfeld2018} the authors propose a trainable activation function in terms of a linear combination of $n$ different, predefined and fixed functions such as hyperbolic tangent (tanh), ReLU and ELU. The added parameters are the $n$ coefficients of the linear combination for each hidden neuron.  
A similar approach is proposed in  \citep{harmon2017} where the authors model the trainable activation function as a linear combination of a predefined set of $n$ normalized fixed activation functions. The added parameters are the coefficients of the linear combination and a set of offset parameters, $\eta$ and $\delta$,  which are used to dynamically offset the normalization range for each predefined function.
Moreover, in order to force the network to choose amongst the predefined activation functions, during the learning process it is required than all the coefficients of the linear combination add to one. This then gives rise to another optimization process unrelated to the classic learning procedure for neural networks




\subsection{Other approaches}
\label{subsec:otherApproaches}
Two interesting and successful approaches are Maxout\citep{goodfellow2013,sun2018}
and NIN\citep{lin2013}. However, despite the good performances, both approaches 
move away from the concept of trainable activation function as it has been previously discussed insofar as the adaptable function does not correspond to
the neuron activation function by which the neuron output is computed on the basis of
a scalar value (the neuron input) according to the standard two-stage process.
In fact, in Maxout, instead of computing the input $a_i$ of a neuron $i$ and then assigning it as input to a trainable activation function, $n$ input $a_{ij}$ are computed, with $j=1,\dots,n$, by $n$ trainable linear functions, and then the maximum is taken over the output of these linear functions. NIN instead represents an approach used specifically in the case of convolutional neural networks, wherein the nonlinear parts of the filters are replaced with
a fully connected neural network acting on all channels simultaneously. 

Another interesting way to tackle the problem is to use interpolating functions as in
\citep{scardapane2016,trentin2001}. For example, in \citep{scardapane2016} the authors propose an adaptable activation function by using a cubic spline interpolation, whose $q$ control points for each neuron are adapted in the optimization phase. External methods to classic approaches in neural networks are needed to train the added parameters $q*m$, where $m$ is the number of hidden neurons.
\subsection{Summarizing}
\label{subsec:summarizinApproaches}
In all the known approaches, to the best of the our knowledge, either the expressive power of the trainable activation functions is limited or they add new learning mechanisms, constraints and categories of parameters, by contrast in our approach we achieve a feed-forward neural network with \textit{trainable} activation functions by a feed-forward neural network with \textit{fixed} activation functions, thus leaving unaltered the classic learning mechanisms and categories of parameters. 
Thus, in our approach a number of attractive properties are simultaneously satisfied:
\textit{p1}) the trainable activation function can approximate arbitrarily well any continuous one-to-one mapping defined on a compact input domain, \textit{p2}) any standard learning mechanism for neural network can be directly and easily applied, \textit{p3}) no learning process in addition to those classically used for neural networks is added, \textit{p4}) the added parameters are network weights or biases , therefore any classical regularization method can be used, including the possibility of imposing sparsity by using norms such as $l_1$.  

None of the known approaches possess all these properties simultaneously.  For example, property \textit{p1} is non satisfied for all the approaches discussed in Section \ref{subsec:parameterizedSAF} and \ref{subsec:ensemble}, the approaches discusses in Section \ref{subsec:linearCombinationBF} either do no satisfy property \textit{p1} as in \citep{agostinelli2014} or property \textit{p3} as in \citep{scardapane2018}.    

Interestingly, as we will discuss in Section \ref{sec:systemArch} our architecture represents a general framework in which all the approaches described in Section \ref{subsec:linearCombinationBF} and some of the approaches in Section \ref{subsec:ensemble} can be included, insofar as any linear combination of $m$ one-variable functions can be represented by a sub-network with $m$ hidden neurons. 

\section{System architecture}
\label{sec:systemArch}
\subsection{Proposed model: Variable Activation Function Subnetwork}
In general, as already introduced in Section \ref{sec:introduction}, in a MLFF network the output of a neuron $i$  belonging to the $l$-th layer is obtained by a two-step computation (see \citep{bishop2006}, Chapter 4). The first step computes the input $a^l_i=\sum\limits_j w^{l}_{ij} z^{l-1}_j + b^l_i$, where $j$ runs over neuron's indexes (or network input values) of the previous layer $l-1$, which send connections to $i$, $z_j^{l-1}$ are the output  of the neurons belonging to the previous layer (or network input values),  $w_{ij}^l$ are the connection weights going from the neurons $j$ of the previous layer $l-1$ to the neuron $i$, and $b^l_i$  is the bias associated to the neuron $i$ . 
The output of the neuron $i$ is then computed in a second step transforming the input $a_i^l$ using a fixed activation function $f$, obtaining $z^l_i=f(a^l_i)$.

The key idea of our approach is to implement the second step of the computation by a ``small'' one-hidden layer sub-network, with $k$ hidden neurons and just one input and one output neuron.  Let us call it \textit{Variable Activation Function} (henceforth, VAF) sub-network. So, a VAF for a neuron $i$ can be described as a network composed by:
\begin{itemize}
    \item an hidden layer, composed by $k>1$ neurons directly connected to the neuron $i$ by a set of weights $\alpha_h$ , with $h=1,2,\cdots,k$;
    \item a fixed activation function $g$ for the $k$ hidden neurons; 
    \item an output layer composed by a single neuron connected to all the neurons of the hidden layer by a set of weights $\beta_h$, with $h=1,2,\cdots k$.
\end{itemize}
The computation of a VAF sub-network associated to a neuron $i$ can be described as follows: VAF sub-network is fed with the input $a_i$ of the neuron $i$, then the $k$ neurons of the hidden layer compute $k$ outputs as $y_h = g(\alpha_h a_i + \alpha_{0h})$ with $h=1,2, \cdots,k$, while the output neuron computes $z=\sum_h \beta_h y_h + \beta_{0}$. $\alpha_{h}$ and $\alpha_{0h}$ are weights and biases of the hidden layer of the VAF sub-network, respectively, and  $\beta_h$ and $\beta_{0}$ are weights and bias of the output layer of VAF sub-network, respectively. In this way the output $z_i$ of the neuron $i$ can be expressed as:
%
%
\begin{equation}\label{eq:vafGeneralEq}
z_i=VAF(a_i)= \sum\limits_{j=1}^k \beta_j g(\alpha_j a_i + \alpha_{0j})+\beta_0
\end{equation}
$\alpha_j$, $\alpha_{0j}$, $\beta_j$ and $\beta_0$  are the parameters to be learned from data during the training process.

\begin{figure}[t]
\begin{center}
    \begin{tikzpicture}
[   cnode/.style={draw=black,fill=#1,minimum width=3mm,circle},
]
\node[line width=0.1mm, font=\fontsize{0.1}{0.1}\selectfont, circle split, draw, minimum size=0.1cm,rotate=90] (s) at (6,-3) {\scalebox{1.0}{\rotatebox{-90}{$\Sigma$}} \nodepart{lower} \scalebox{1}{\rotatebox{-90}{$I$}}};

\node[line width=0.1mm, font=\fontsize{0.1}{0.1}\selectfont, circle split, draw, minimum size=0.1cm,rotate=90] (x) at (0,-3) {\scalebox{1}{\rotatebox{-90}{$x$}} \nodepart{lower} \scalebox{1}{\rotatebox{-90}{$I$}}};
    \foreach \x in {1,...,4}
    {   \pgfmathparse{\x<4 ? \x : "n"}
        
        \node[line width=0.1mm, font=\fontsize{5}{5}\selectfont, circle split, draw, minimum size=0.1cm,rotate=90] (p-\x) at (3,{-\x-div(\x,4)}) {\rotatebox{-90}{$\cdot$} \nodepart{lower} \scalebox{0.7}{$ReLU$}};
        \draw (p-\x) -- node[above,sloped,pos=0.35] {\scalebox{0.5}{$\beta_\x$}} (s);
    }
    
  \foreach \y in {1,...,4}
        { \pgfmathparse{\y<4 ? \y : "n"}  
        	\draw (x) -- node[above,sloped,pos=0.57] {\scalebox{0.5}{$\alpha_\y$}} (p-\y);
        }
    \node at (3,-4) {$\vdots$};
++\end{tikzpicture}
    \caption{A general VAF scheme; with $I$ we indicate the identity function}
    \label{fig:vaf1scheme}
\end{center}
\end{figure}
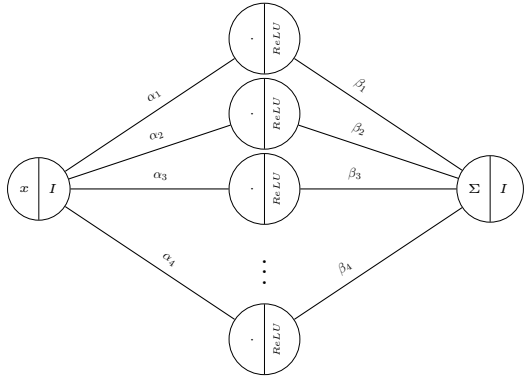

A general schema of a VAF unit is shown in figure \ref{fig:vaf1scheme}.
This schema enables one to approximate arbitrarily well any activation function provided that:
\begin{itemize}
\item the number $k$ of hidden neurons in the VAF is sufficiently large;
\item the activation function $g$ of the hidden layer is a not-polynomial function.
\end{itemize}
As already discussed in Section \ref{sec:introduction}, the first condition is given in \citep{hornik1989,hornik1991}, where it was shown that a shallow networks can approximate any continuous function provided that a sufficient number of hidden neurons are available and that the activation function is continuous, bounded and non-constant. 
This result was generalized in \citep{leshno1993}, where it is proved that a shallow network can approximate any continuous function to any degree of accuracy if and only if the network's activation function is not polynomial.
Therefore a VAF activation function can substitute any other network activation function without loss in generality, and having as overhead only an increase in the number of networks parameters, that is equal to $N \cdot (3k+1)$ with $N$ total number of the hidden neurons of the network.
Anyway, the number of required parameters can drop to $L \cdot (3k+1)$, with $L$ number of hidden layers, if we adopt the \textit{shared weights principle}, so that the functions on the same layer share the same VAF weights. 
With this design choice, we reduce the number of parameters by making the reasonable assumption that if one function is good for a single neuron, then it should also be good for the other neurons of the same layer.
This assumption can also be motivated, instead of under the profile of the sub-networks weights, in terms of activation function of a classic neural networks used in the neural network literature, where neurons on to same layer exhibit the same activation function. Summing up, under the shared weights principle for every network layer the only added hyper-parameters to set are: 
\begin{itemize}
    \item the number $k$ of hidden neurons of the VAF subnetwork;
    \item the activation function $g(\cdot)$ of the VAF hidden neurons.
\end{itemize}

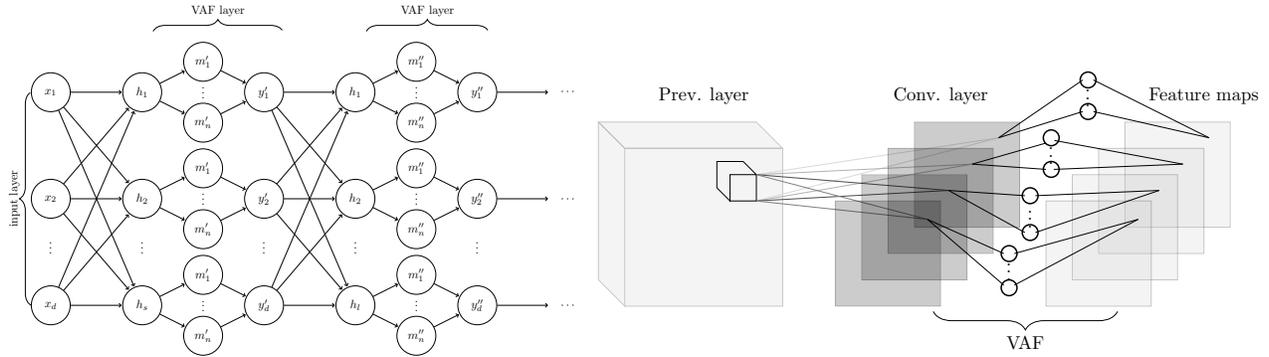
\begin{figure}[t]
    \scalebox{0.45}{\begin{tikzpicture}[shorten >=1pt,scale=0.9]
		\tikzstyle{unit}=[draw,shape=circle,minimum size=1.15cm]

		\node[unit](i1) at (-3,3.5){$x_1$};
		\node[unit](i2) at (-3,0){$x_2$};
		\node(dots) at (-3,-1.5){\vdots};
		\node[unit](id) at (-3,-3.5){$x_d$};

		\node[unit](x1) at (0,3.5){$h_1$};
		\node[unit](x2) at (0,0){$h_2$};
		\node(dots) at (0,-1.5){\vdots};
		\node[unit](xd) at (0,-3.5){$h_s$};
		
		\node[unit](m11) at (2,4.5){$m'_1$};
		\node(dots) at (2,3.6){\vdots};
		\node[unit](mn1) at (2,2.5){$m'_n$};

		\node[unit](m12) at (2,1){$m'_1$};
		\node(dots) at (2,0.1){\vdots};
		\node[unit](mn2) at (2,-1){$m'_n$};		

		\node[unit](m1d) at (2,-2.5){$m'_1$};
		\node(dots) at (2,-3.4){\vdots};
		\node[unit](mnd) at (2,-4.5){$m'_n$};

		\node[unit](y1) at (4,3.5){$y'_1$};
		\node[unit](y2) at (4,0){$y'_2$};
		\node(dots) at (4,-1.5){\vdots};
		\node[unit](yd) at (4,-3.5){$y'_d$};

		\node[unit](h1) at (7,3.5){$h_1$};
		\node[unit](h2) at (7,0){$h_2$};
		\node(dots) at (7,-1.5){\vdots};
		\node[unit](hl) at (7,-3.5){$h_l$};

		\node[unit](m11b) at (9,4.5){$m''_1$};
		\node(dots) at (9,3.6){\vdots};
		\node[unit](mn1b) at (9,2.5){$m''_n$};
				
		\node[unit](m12b) at (9,1){$m''_1$};
		\node(dots) at (9,0.1){\vdots};
		\node[unit](mn2b) at (9,-1){$m''_n$};		
		
		\node[unit](m1db) at (9,-2.5){$m''_1$};
		\node(dots) at (9,-3.4){\vdots};
		\node[unit](mndb) at (9,-4.5){$m''_n$};								

		\node[unit](y1b) at (11,3.5){$y''_1$};
		\node[unit](y2b) at (11,0){$y''_2$};
		\node(dots) at (11,-1.5){\vdots};
		\node[unit](ylb) at (11,-3.5){$y''_d$};
		
		\node[unit,draw=none] (inv1) at (14,3.5) {$\cdots$};
		\node[unit,draw=none] (inv2) at (14,0) {$\cdots$};			
		\node[unit,draw=none] (inv3) at (14,-3.5) {$\cdots$};

		\draw[->] (i1) -- (x1);
		\draw[->] (i1) -- (x2);
		\draw[->] (i1) -- (xd);		
		\draw[->] (i2) -- (x1);
		\draw[->] (i2) -- (x2);
		\draw[->] (i2) -- (xd);		
		\draw[->] (id) -- (x1);
		\draw[->] (id) -- (x2);
		\draw[->] (id) -- (xd);

		\draw[->] (x1) -- (m11);
		\draw[->] (x1) -- (mn1);
		
		\draw[->] (x2) -- (m12);
		\draw[->] (x2) -- (mn2);
		
		\draw[->] (xd) -- (m1d);
		\draw[->] (xd) -- (mnd);

		\draw[->] (m11) -- (y1);
		\draw[->] (mn1) -- (y1);
		\draw[->] (m12) -- (y2);
		\draw[->] (mn2) -- (y2);		
		\draw[->] (m1d) -- (yd);
		\draw[->] (mnd) -- (yd);		

		\draw[->] (y1) -- (h1);
		\draw[->] (y1) -- (h2);
		\draw[->] (y1) -- (hl);		
		\draw[->] (y2) -- (h1);
		\draw[->] (y2) -- (h2);
		\draw[->] (y2) -- (hl);		
		\draw[->] (yd) -- (h1);
		\draw[->] (yd) -- (h2);
		\draw[->] (yd) -- (hl);

		\draw[->] (h1) -- (m11b);
		\draw[->] (h1) -- (mn1b);
		
		\draw[->] (h2) -- (m12b);
		\draw[->] (h2) -- (mn2b);
		
		\draw[->] (hl) -- (m1db);
		\draw[->] (hl) -- (mndb);

		\draw[->] (m11b) -- (y1b);
		\draw[->] (mn1b) -- (y1b);
		\draw[->] (m12b) -- (y2b);
		\draw[->] (mn2b) -- (y2b);		
		\draw[->] (m1db) -- (ylb);
		\draw[->] (mndb) -- (ylb);				

		\draw[->] (y1b) -- (inv1);
		\draw[->] (y2b) -- (inv2);
		\draw[->] (ylb) -- (inv3);

		\draw [decorate,decoration={brace,amplitude=10pt},xshift=-4pt,yshift=0pt] (-3.5,-3.5) -- (-3.5,3.5) node [black,midway,rotate=90,xshift=0cm,yshift=0.5cm]{input layer};
		\draw [decorate,decoration={brace,amplitude=10pt},xshift=-4pt,yshift=0pt] (0.5,5.5) -- (4.75,5.5) node [black,midway,yshift=+0.6cm]{VAF layer};
		\draw
		[decorate,decoration={brace,amplitude=10pt},xshift=-4pt,yshift=0pt] (7.5,5.5) -- (11.5,5.5) node [black,midway,yshift=+0.6cm]{VAF layer};		
		\end{tikzpicture}}
    \scalebox{0.7}{\begin{tikzpicture}
	
    \draw [black,  thick] (7.3,0.35) ellipse [x radius = 0.15cm, y radius = 0.15cm];
	\node at (7.3,0.7){\begin{tabular}{c}$\vdots$\\ \end{tabular}};
	
    \draw [black,  thick] (7.3,1.0) ellipse [x radius = 0.15cm, y radius = 0.15cm];

    \draw [black,  thick] (7.7,1.4) ellipse [x radius = 0.15cm, y radius = 0.15cm];
    \node at (7.7,1.7){\begin{tabular}{c}$\vdots$\\ \end{tabular}};
    
    \draw [black,  thick] (7.7,2.1) ellipse [x radius = 0.15cm, y radius = 0.15cm];

    \draw [black,  thick] (8.1,2.6) ellipse [x radius = 0.15cm, y radius = 0.15cm];
    \node at (8.1,2.9){\begin{tabular}{c}$\vdots$\\ \end{tabular}};
    
    \draw [black,  thick] (8.1,3.2) ellipse [x radius = 0.15cm, y radius = 0.15cm];

	\draw [black,  thick] (8.8,3.7) ellipse [x radius = 0.15cm, y radius = 0.15cm];
	\node at (8.8,4.0){\begin{tabular}{c}$\vdots$\\ \end{tabular}};
	
	\draw [black,  thick] (8.8,4.3) ellipse [x radius = 0.15cm, y radius = 0.15cm];       

	\node at (1.5,4){\begin{tabular}{c}Prev. layer\\ \end{tabular}};
	\draw[fill=green,opacity=0.2,draw=black] (0,0) -- (3,0) -- (3,3) -- (0,3) -- (0,0)--(-0.5,+0.5)--(-0.5,3.5)--(0,3)--(3,3)--(2.5,3.5)--(-0.5,3.5)--(0,3);	
	\node at (6,4){\begin{tabular}{c}Conv. layer\\ \end{tabular}};	
	\draw (2,2) -- (2.5,2) -- (2.5,2.5) -- (2,2.5) -- (2,2)--(1.75,2.25)--(1.75,2.75)--(2.25,2.75)--(2.5,2.5);
.
	\node at (11,4){\begin{tabular}{c}Feature maps\\ \end{tabular}};	

	\draw[color=red] (2.5,2.5) edge (7.1,3.2);
	\draw[color=red] ((2.5,2) -- (7.1,3.2);
	\draw[color=blue] (2.5,2.5) -- (6.6,2.7);
	\draw[color=blue] ((2.5,2) -- (6.6,2.7);
	\draw[color=brown] (2.5,2.5) -- (6.15,2.2);
	\draw[color=brown] ((2.5,2) -- (6.15,2.2);	
	\draw[color=violet](2.5,2.5) -- (5.75,1.65);
	\draw[color=violet] ((2.5,2) -- (5.75,1.65);						
	\node at (5.5,4){\begin{tabular}{c}\end{tabular}};
	\node at (6.3,0){\begin{tabular}{c}\end{tabular}};
	\node at (6.8,0.5){\begin{tabular}{c}\end{tabular}};	
	\node at (7.3,1){\begin{tabular}{c}\end{tabular}};
	\node at (7.8,1.5){\begin{tabular}{c}\end{tabular}};	
	
	\draw[fill=black,opacity=0.2,draw=black] (5.5,1.5) -- (7.5,1.5) -- (7.5,3.5) -- (5.5,3.5) -- (5.5,1.5);
	\draw[fill=black,opacity=0.2,draw=black] (5,1) -- (7,1) -- (7,3) -- (5,3) -- (5,1);
	\draw[fill=black,opacity=0.2,draw=black] (4.5,0.5) -- (6.5,0.5) -- (6.5,2.5) -- (4.5,2.5) -- (4.5,0.5);
	\draw[fill=black,opacity=0.2,draw=black] (4,0) -- (6,0) -- (6,2) -- (4,2) -- (4,0);
	
	\draw[fill=green,opacity=0.2,draw=black] (9.5,1.5) -- (11.5,1.5) -- (11.5,3.5) -- (9.5,3.5) -- (9.5,1.5);

	\draw[fill=green,opacity=0.2,draw=black] (9,1) -- (11,1) -- (11,3) -- (9,3) -- (9,1);
	\draw[fill=green,opacity=0.2,draw=black] (8.5,0.5) -- (10.5,0.5) -- (10.5,2.5) -- (8.5,2.5) -- (8.5,0.5);
	\draw[fill=green,opacity=0.2,draw=black] (8,0) -- (10,0) -- (10,2) -- (8,2) -- (8,0);

	\draw[color=black] (5.75,1.65) edge node[above=-0.1,text=violet] {} (7.2,0.4);		
	\draw[color=black] (5.75,1.65) edge node[above=-0.1,text=violet ]{} (7.2,1.0);
	\draw[color=black] (7.4,0.4) edge node[above=-0.1,text=violet ]{} (9.75,1.65);
	\draw[color=black] (7.4,1.0) edge node[above=-0.1,text=violet ]{} (9.75,1.65);

	\draw[color=black] (6.15,2.2) edge node[above=-0.1,text=brown ]{} (7.6,2.1);
	\draw[color=black] (6.15,2.2) edge node[above=-0.1,text=brown ]{} (7.6,1.4);
	\draw[color=black] (7.8,2.1) edge node[above=-0.1,text=brown ]{} (10.15,2.2);
	\draw[color=black] (7.8,1.4) edge node[above=-0.1,text=brown ]{} (10.15,2.2) (10.15,2.2);

	\draw[color=black] (6.6,2.7) edge node[above=-0.1,text=blue]{} (8,2.6);	
	\draw[color=black] (6.6,2.7) edge node[above=-0.1,text=blue]{} (8,3.1);						
	\draw[color=black] (8.15,2.6) edge node[above=-0.1,text=blue]{} (10.6,2.7);	
	\draw[color=black] (8.16,3.15) edge node[above=-0.1,text=blue]{} (10.6,2.7);		
	
	\draw[color=black]  (7.1,3.2) edge node[above=-0.1,text=red]{} (8.7,3.7);	
	\draw[color=black]  (7.1,3.2) edge node[above=-0.1,text=red]{} (8.7,4.3);	
	\draw[color=black]  (8.9,3.7) edge node[above=-0.1,text=red]{} (11.1,3.2);	
	\draw[color=black]  (8.9,4.3) edge node[above=-0.1,text=red]{} (11.1,3.2);	
	
	\node at (5.3,0.2){\begin{tabular}{c}\end{tabular}};
	\node at (5.7,0.7){\begin{tabular}{c}\end{tabular}};	
	\node at (6.2,1.2){\begin{tabular}{c}\end{tabular}};	
	\node at (6.7,1.7){\begin{tabular}{c}\end{tabular}};	
		
	\node at (9.3,0.2){\begin{tabular}{c}\end{tabular}};
	\node at (9.7,0.7){\begin{tabular}{c}\end{tabular}};	
	\node at (10.2,1.2){\begin{tabular}{c}\end{tabular}};	
	\node at (10.7,1.7){\begin{tabular}{c}\end{tabular}};			

	\draw [decorate,decoration={brace,amplitude=10pt,mirror},xshift=-4pt,yshift=0pt] (6,-0.1) -- (9.5,-0.1) node [black,midway,yshift=-0.6cm]{VAF};

	\end{tikzpicture}    }
    \caption{an example of VAF in a 2 full connected network (on the left) and in a convolutional layer (on the right)}
    \label{fig:vaf23scheme}
\end{figure}

It is worth to emphasize the fact that, in our approach, we have a neural network architecture which is still a MLFF network with fixed activation functions, without adding any external structure or parameters. Let us clarify this aspect (see also Figure \ref{fig:vaf23scheme}). Given a neuron $i$ belonging to $l$-th layer of a MLFF network $Net$, its output is computed as $z^l_i=f(a^l_i)$, in our approach we substitute the activation function $f$ with the Identity function, thus obtaining $z^l_i=a^l_i$. Then, we add a VAF sub-network which receives as input variable the output $z^l_i$ of the neuron $i$ and computes its output as defined in eq. \ref{eq:vafGeneralEq}. Finally, this output is sent as input of the next layer $l+1$ of $Net$. This procedure is uniformly performed for all the neurons of the MLFF network $Net$, but the output layer.  Thus, one obtains a new neural network $VafNet$ which is still a MLFF network with fixed activation functions, however it behaviours as $Net$ equipped with trainable activation functions expressed in terms of eq. \ref{eq:vafGeneralEq}.  
Consequently, any standard training procedure can be left unaltered (e.g., Stochastic Gradient Descent).

Figure \ref{fig:vaf23scheme} shows how a VAF network can be integrated into a common multilayer full-connected neural network (on the left) and in a convolutional neural network (on the right).

Notably, given that a VAF subnetwok performs a linear combination of one-variable functions, any approach discussed in Section \ref{subsec:linearCombinationBF} can be included in this schema, provided to choose suitably the activation function $g$ and the parameters $\alpha$ and $\beta$.   

\subsection{VAF network learning}
\label{subsec:vafLearning}
As discussed above, our neural architecture including VAF is a MLFF network, consequently it can be trained using any learning algorithm dedicated to MLFF network. 
However, in case of the same VAF acting uniformly for all neurons of a layer, then there is the constrain that the weights of VAF networks should be considered \textit{shared weights}. 
From an implementation point of view this corresponds to consider a VAF network as a function convolving with the $a_i^l$ values \citep{lin2013}. 
The weight values of the VAF, being few and connected to each unit, influence the behaviour of the entire network, therefore their behaviour must be taken into consideration during the training phase, and, in particular, the initial value of the VAF weights can be decisive.
The training of a neural network usually starts initializing the weights and biases in a random way \citep{bishop2006}, or using any initialization rule as for example \citep{glorot2010}. 
Although these approaches can also be followed in our case, it is possible to choose different solutions for the VAF weights initialization. 
In particular, a possible alternative is to select the initial weights of the VAF so that at the beginning of the learning process the VAF networks approximate a fixed function. 
For example, we can select a classic activation function as ReLU or sigmoid, or the $f$ activation function associated to the other hidden layers of the network. 
In this way hypothetically the function would start from a notoriously already valid form in which the training process should only modify it just enough to improve the performance of the network based on the training data.
However, it should be kept in mind that this choice can affect negatively the solution generated by the learning process, given that the resulting VAF can be too similar to the initial function.

\begin{algorithm}[H]
\SetAlgoLined
\KwIn{$TS$, $VS$, $net$, $MaxEpochs$: $TS$ and $VS$ are training and validation datasets, respectively; $net$ is the network to be trained; $MaxEpochs$ is the maximum number of epochs}
\KwOut{$trainNet$: Trained net }
 $net \gets weightAndBiasInitialization(net)$ \;
 $bestNet \gets net$ \;
 
 $n \gets 0$, $minErr \gets MAX$ \;
 \Repeat{$n > MaxEpochs$ OR earlyStoppingCriteria(errorT,errorV)}{
 	$n \gets n+1$ \;
  	$net  \gets learningAlgorithm(net,TS)$ \;
    $errorT(n) \gets Sim(net,TS)$\;
    $errorV(n) \gets Sim(net,VS)$\;
    \If {$errorV(n) < minErr$} {
    	$minErr \gets errorV(n)$\;
        $bestNet \gets net$\;
    }
 }
 $trainNet \gets bestNet$\;

 \caption{Standard learning schema}
 \label{algo:standard-learning}
\end{algorithm}

\section{Experimental results}
\label{sec:expFramework}
In this section we provide an experimental evaluation of the proposed trainable activation function architecture. In order to achieve a first clue on the validity of our approach, and some heuristic indications for the initialization strategies of VAF networks, in Section \ref{subsec:preliminaryExp} we report some preliminary experiments on \textit{Sensorless}, a relatively small classification dataset used as standard benchmark for supervised techniques.

On the basis of the results of these experiments, we performed two different series of experiments to test our approach on MLFF networks. In the former, we consider standard MLFF networks (Section \ref{subsubsection:firstExp}) , and in the latter convolutional MLFF networks (Section \ref{subsubsection:secondExp}). In Section \ref{subsubsection:firstExp} we consider both classification and regression problems using $20$ different datasets. In Section \ref{subsubsection:secondExp} we consider more large-scale dataset as MNIST, Fashion MNIST and CIFAR10.

\subsection{VAF subnetworks: Activation functions, number of hidden neurons and weight initialization}
\label{subsec:preliminaryExp}
For a preliminary analysis of the validity of our approach, and for defining some heuristic choices about VAF subnets such as the number and the activation functions of the hidden neurons, we perform a series of experiments on \textit{Sensorless} dataset (see table \ref{table:dataset} for details), partitioning it in a random sample of 60\% for training, 20\% for validation and another 20\% for testing.  
According to what was also reported in \citep{scardapane2018}, if one uses a standard shallow network, i.e., $1$-hidden layer network, we found that tanh is the best fixed activation function for this dataset. In particular, using a shallow network with $50$ hidden neurons we obtained an accuracy on the test set very close to 100\%.  Thus, to better investigate the impact of our approach we chose a more ``difficult situation'' for a shallow network using network models with a small number of hidden neurons. More in detail, we selected three small shallow nets with $5$, $10$ and $20$ hidden neurons.

For each model, We perform a set of experiments using different activation functions. 

Firstly, we train these small networks using as fixed activation functions either $tanh$ or $ReLU$, then we repeat the same experiments substituting the fixed activation functions with VAF subnets as described in Section \ref{sec:systemArch}. We considered several scenarios: 1) different number $k$ of VAF hidden neurons, in particular $k \in \{3,5,7,9,11,15\}$; 2) tanh and ReLU as activation functions for VAF hidden neurons; 3) two different strategies for weight initialization of VAF subnets,  both a classic random initialization and a weight initialization by which VAF subnets have a behaviour very similar to activation functions of the VAF hidden neurons, we will call the latter \textit{specific weight initialization}; 4) as discussed in Section \ref{sec:systemArch}, we examine both the case in which VAF subnets on the same layer share the weights (shared weights principle) and the case in which VAF subnets on the same layer can have different weights.

We trained all the networks using ADAM algorithm \citep{kingma2014} for $500$ epochs. Furthermore, we repeat our experiments for $10$ times. 


\subsubsection*{Results}
\begin{figure}
    \centering
    \begin{subfigure}[b]{0.48\textwidth}
        \includegraphics[width=\textwidth]{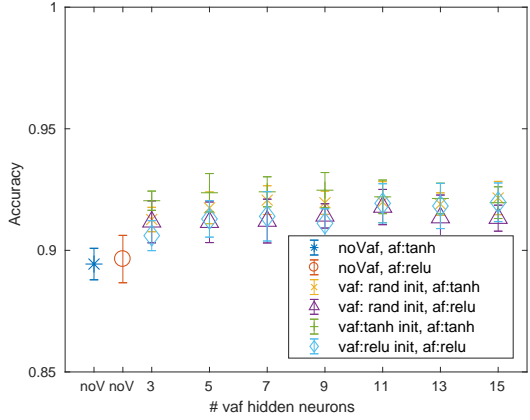}
        \caption{Shallow Network with $5$ hidden neurons}
        \label{subfig:preExpMultipleVAF_hd5}
    \end{subfigure}
    \begin{subfigure}[b]{0.48\textwidth}
        \includegraphics[width=\textwidth]{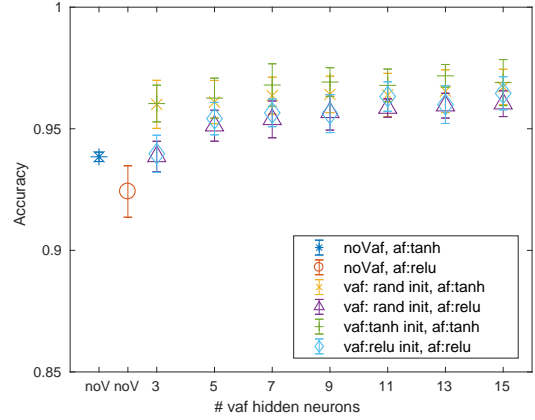}
        \caption{Shallow Network with $10$ hidden neurons}
        \label{subfig:preExpMultipleVAF_hd10}
    \end{subfigure}
    \begin{subfigure}[b]{0.48\textwidth}
        \includegraphics[width=\textwidth]{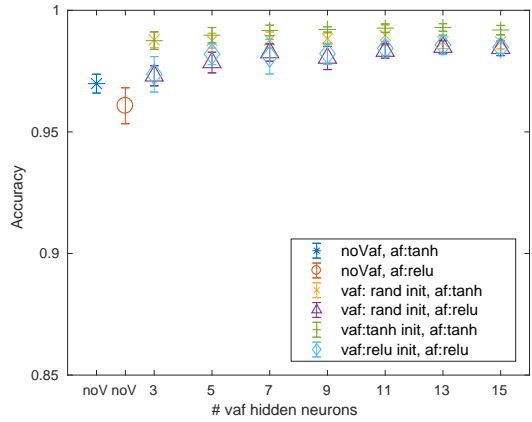}
        \caption{Shallow Network with $20$ hidden neurons}
        \label{subfig:preExpMultipleVAF_hd20}
    \end{subfigure}
   
    \caption{Test accuracy of networks with different VAF subnets on each layer. Using \textit{Sensorless} dataset, we trained three small shallow networks composed of $5$, $10$ and $20$ hidden neurons with fixed activation functions corresponding to either $tanh$ or $ReLU$. In figure such networks are referred as \textit{noVaf}. Then we repeated the same experiments substituting the fixed activation functions with VAF subnets. The number VAF hidden neurons ranges in  $k \in {3,5,7,9,11,15}$, the possible activation functions for VAF hidden neurons are tanh and ReLU . Weight initialization of VAF subnets is  either a classic random initialization or a weight initialization by which VAF subnets have a behaviour very similar to activation functions of the VAF hidden neurons. VAF subnets on the same layer can have different weights.} \label{fig:preExpMultipleVAF}
\end{figure}

\begin{figure}
    \centering
    \begin{subfigure}[b]{0.48\textwidth}
        \includegraphics[width=\textwidth]{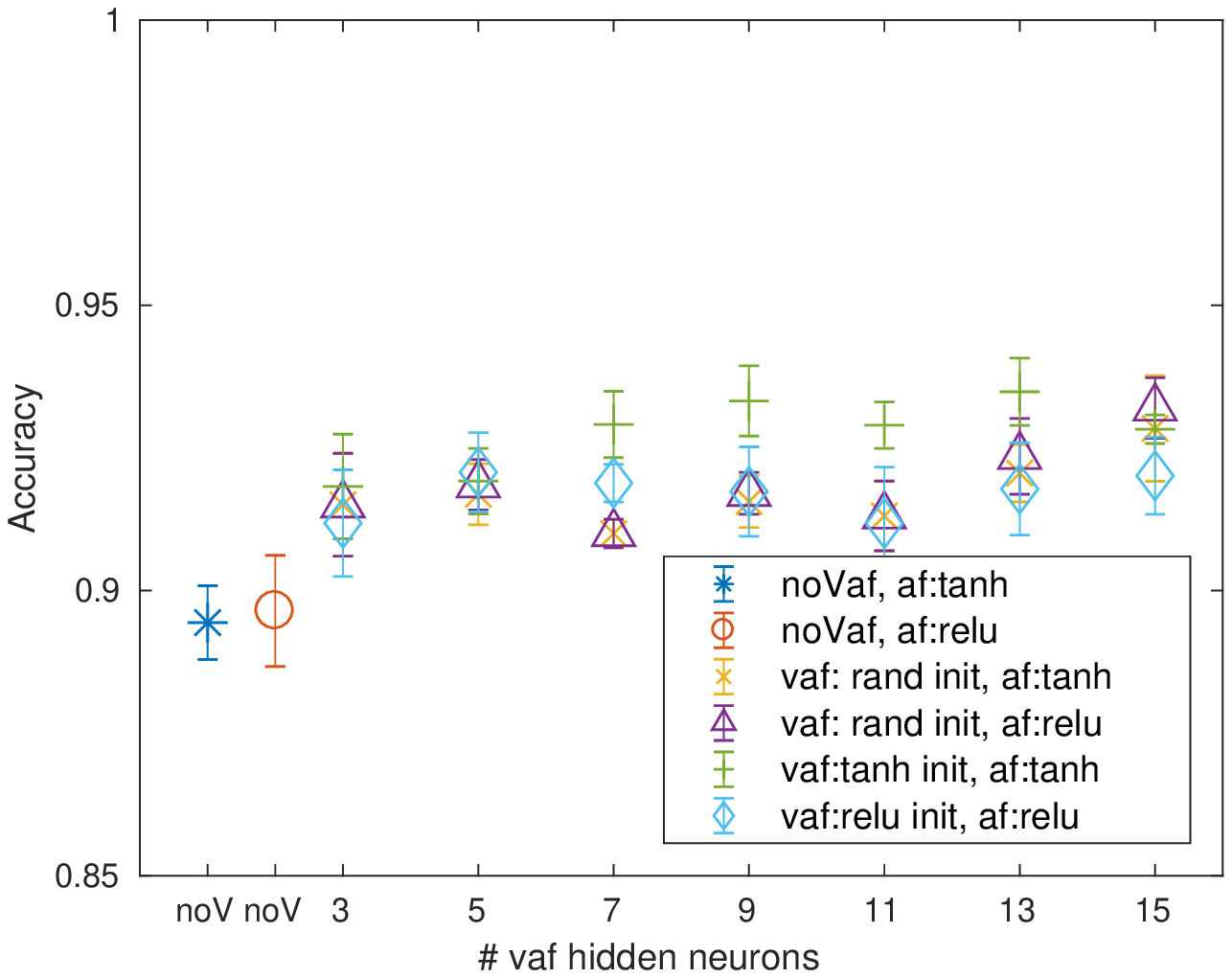}
        \caption{Shallow Network with $5$ hidden neurons}
        \label{subfig:preExpSharedVAF_hd5}
    \end{subfigure}
    \begin{subfigure}[b]{0.48\textwidth}
        \includegraphics[width=\textwidth]{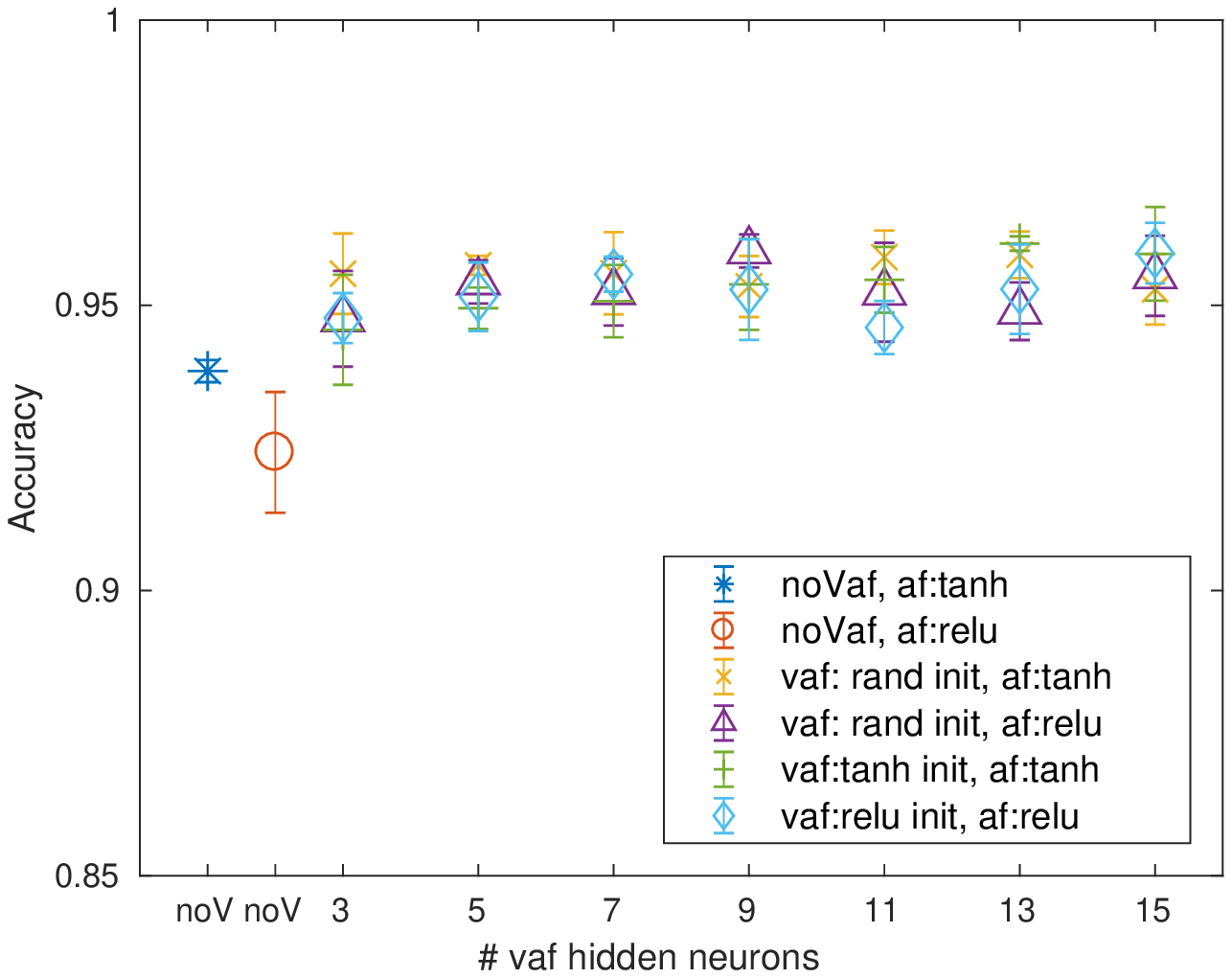}
        \caption{Shallow Network with $10$ hidden neurons}
        \label{subfig:preExpSharedVAF_hd10}
    \end{subfigure}
    \begin{subfigure}[b]{0.48\textwidth}
        \includegraphics[width=\textwidth]{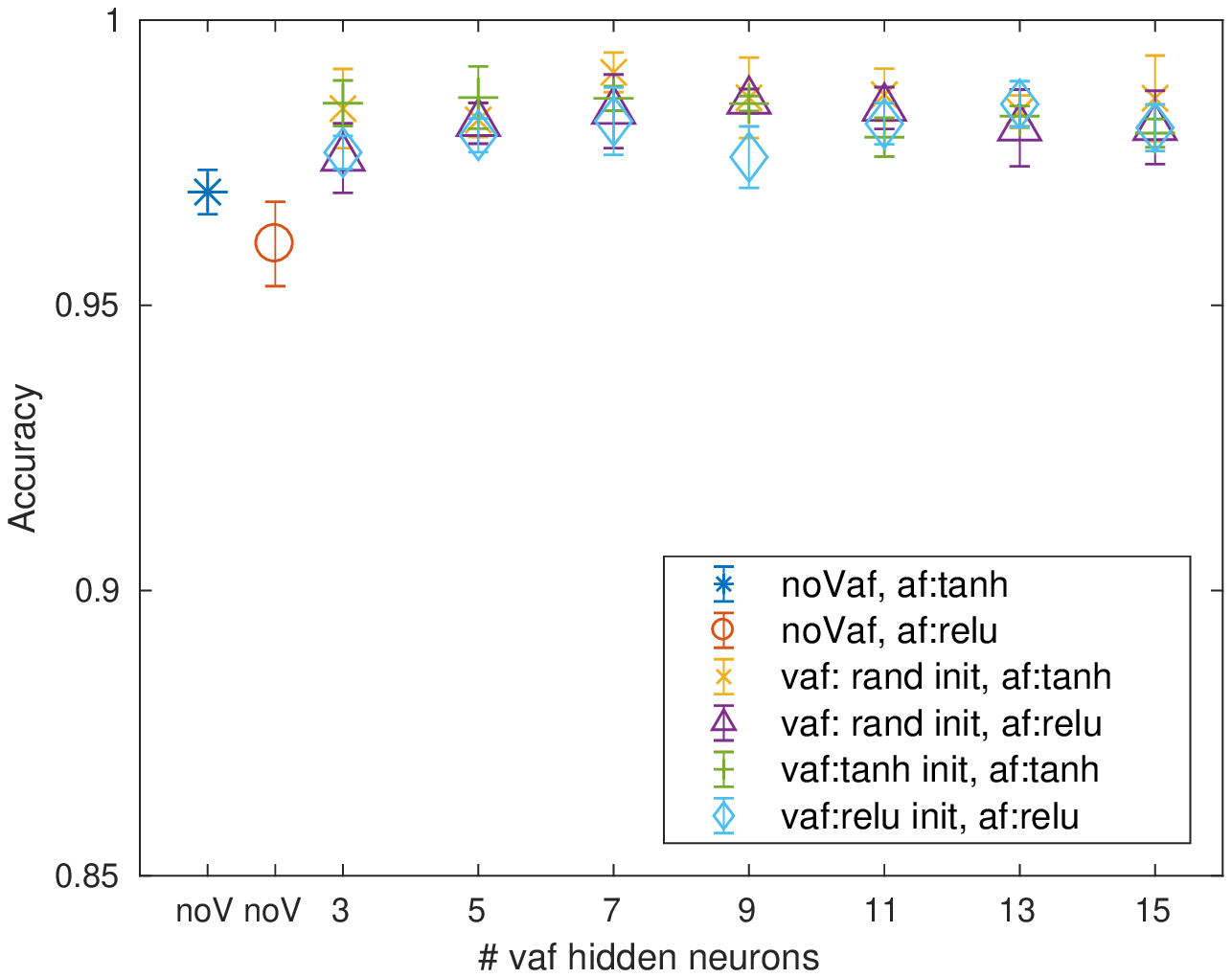}
        \caption{Shallow Network with $20$ hidden neurons}
        \label{subfig:preExpSharedVAF_hd20}
    \end{subfigure}
   
    \caption{Test accuracy of networks with shared VAF subnets on each layer. Using \textit{Sensorless} dataset, we trained three small shallow networks composed of $5$, $10$ and $20$ hidden neurons with fixed activation functions corresponding to either $tanh$ or $ReLU$. In figure such networks are referred as \textit{noVaf}. Then we repeated the same experiments substituting the fixed activation functions with VAF subnets. The number VAF  hidden neurons ranges in  $k \in {3,5,7,9,11,15}$, the possible activation functions for VAF hidden neurons are tanh and ReLU. Weight initialization of VAF subnets is  either a classic random initialization or a weight initialization by which VAF subnets have a behaviour very similar to activation functions of the VAF hidden neurons. VAF subnets on the same layer share the weights.} \label{fig:preExpSharedVAF}
\end{figure}

In Figure \ref{subfig:preExpMultipleVAF_hd5}, \ref{subfig:preExpMultipleVAF_hd10} and \ref{subfig:preExpMultipleVAF_hd20} are reported the results with respect to the shallow networks with $5$, $10$ and $20$ hidden neurons, respectively, in the case in which VAF subnets on the same layer do not share the weights. In Figure \ref{subfig:preExpSharedVAF_hd5}, \ref{subfig:preExpSharedVAF_hd10} and \ref{subfig:preExpSharedVAF_hd20} are reported the results in the case in which VAF subnets on the same layer share the weights. 
 
Notably, one can observe that all the models equipped with VAF subnets outperform the corresponding shallow networks. Interestingly, these results support the possibility of using a shared VAF approach with a fairly low number of VAF hidden neurons, thus having a lower number of parameters to be learned. In fact, the two approaches, non-shared (Figure \ref{fig:preExpMultipleVAF}) and shared (Figure \ref{fig:preExpSharedVAF})  VAF subnets, exhibit a very similar behaviour, and  although in all cases accuracy tends to increase as the number of neurons of the VAF subnets increases, this increase is not always very relevant. The two types of VAF weight initialization seem to give similar results, with slightly better performances for random initialization. 
The use of tanh or ReLU as activation function of  VAF hidden neurons, on the other hand, seems to significantly change the network performance. In fact, the accuracy obtained by networks with  ReLU activation function for the VAF hidden neurons is uniformly lower than those obtained with the tanh activation function. We suppose that this result is due to the fact that we are always using shallow nets.

\begin{figure}
    \centering
    \begin{subfigure}[b]{0.7\textwidth}
        \includegraphics[width=\textwidth]{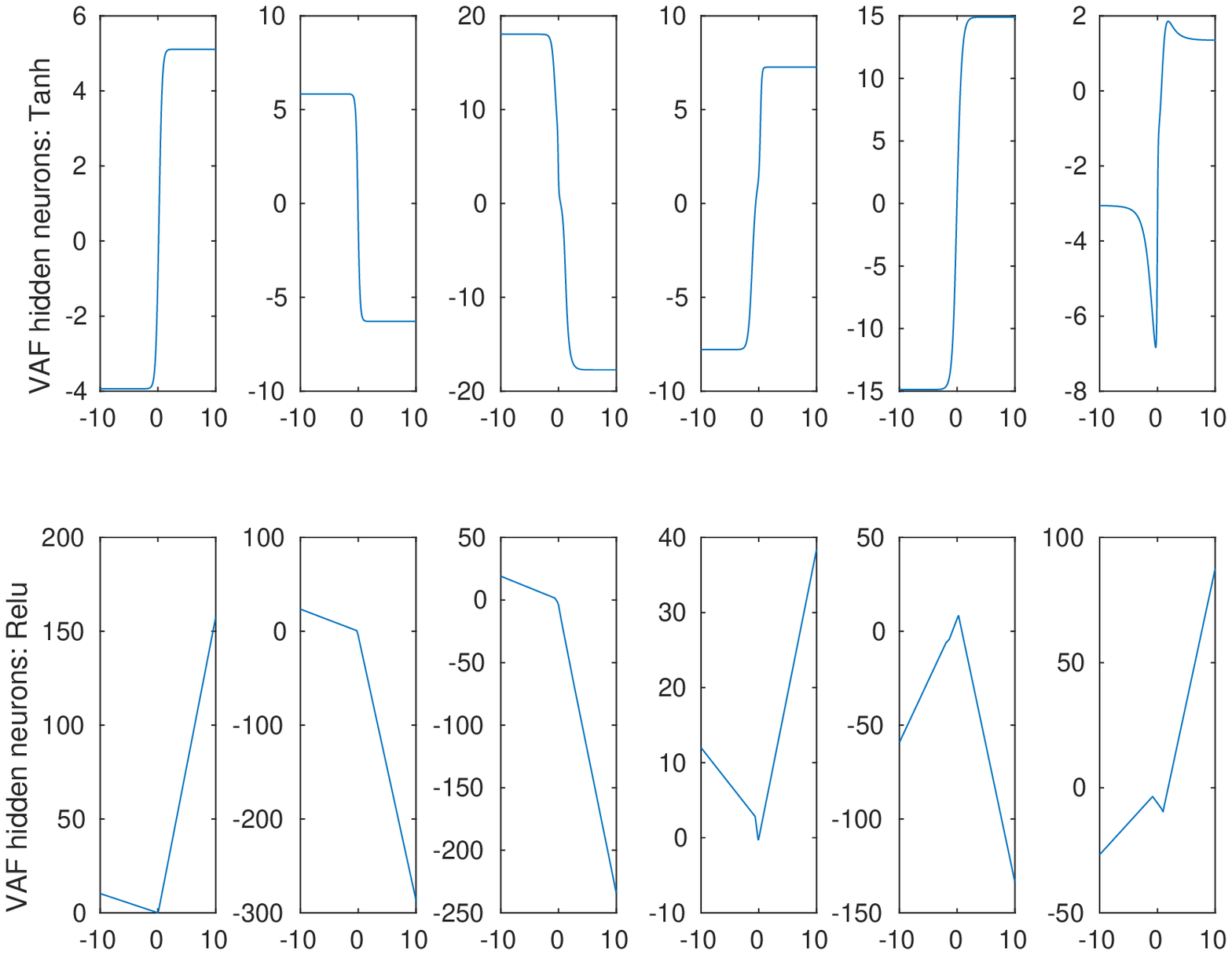}
        \caption{Random Initialization}
        \label{subfig:vafExamplesRandInit}
    \end{subfigure}
    
    \begin{subfigure}[b]{0.7\textwidth}
        \includegraphics[width=\textwidth]{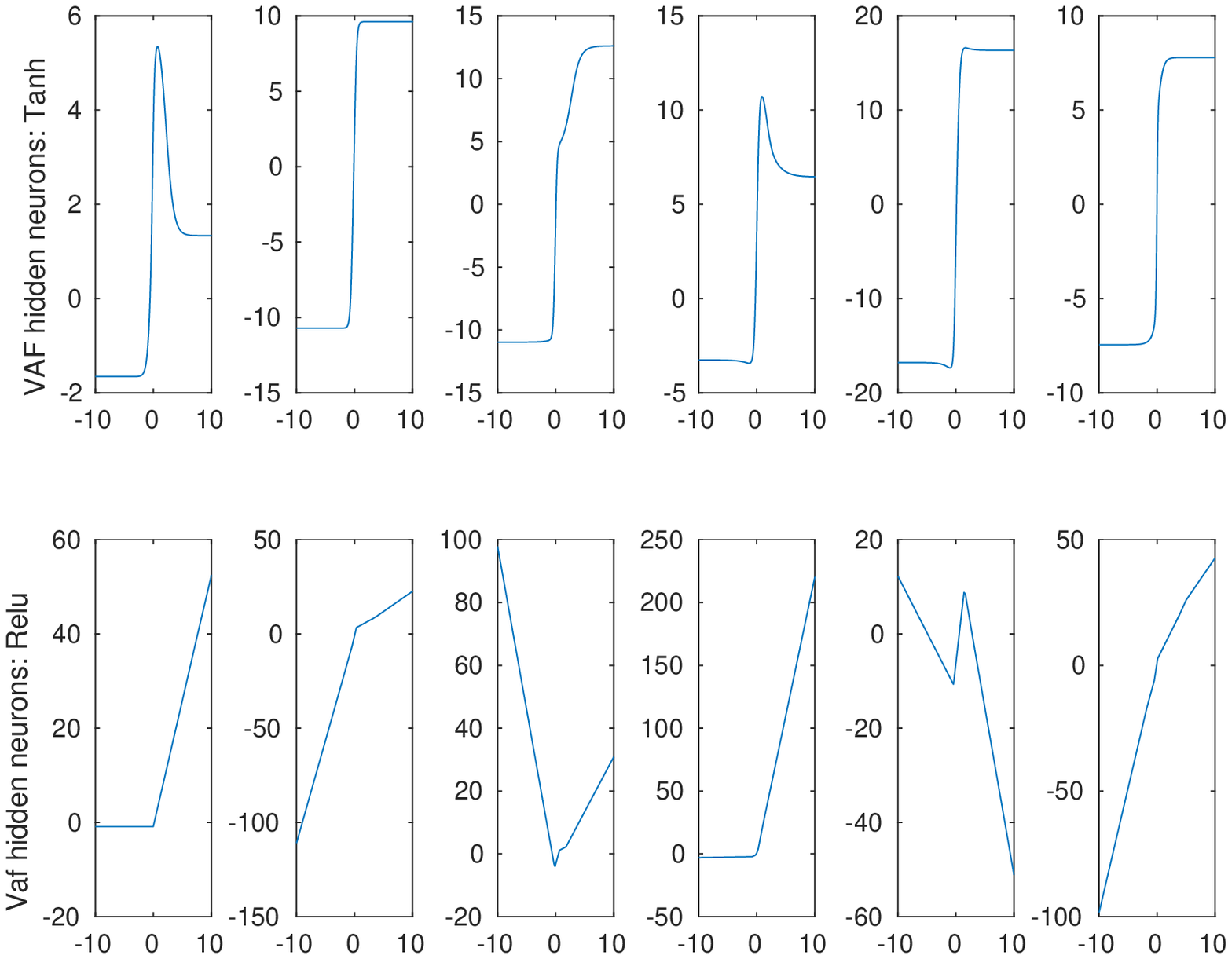}
        \caption{Specific Initialization}
        \label{subfig:vafExamplesSpecInit}
    \end{subfigure}
   
    \caption{Examples of trained VAF subnetworks. On the y-axis we plot the output value of the VAF. In \ref{subfig:vafExamplesRandInit} are plotted trained VAF subnetworks when a random weight initialization is chosen. In \ref{subfig:vafExamplesSpecInit} when a specific weight initialization is chosen}\label{fig:vafExamples}
\end{figure}

In Figure \ref{subfig:vafExamplesRandInit} and \ref{subfig:vafExamplesSpecInit} are reported the output values of trained VAF subnetworks when a random or a specific weight initialization is chosen, respectively. One can note that the resulting activation functions are often strongly different from the classic tanh and ReLU, and that they exhibit similarly a high degree of non-linearity.

\subsection{Full-connected MLFF networks: classification and regression}
\label{subsubsection:firstExp}
In this experimental scenario we focus on evaluating the impact of both VAF subnetworks and VAF weight initialization using full-connected MLFF networks with $1$ or $2$ hidden layers trained on $20$ public datasets (see Table \ref{table:dataset}). $10$ of these datasets are suitable for classification problems, and $10$ for regression problems. 
The number of hidden neurons varies in the set $\{10, 25, 50, 100\}$, but for neural networks with $2$ hidden layers we only selected neural networks with a number of hidden neurons belonging to the  first layer larger than the number of hidden neurons of the second layer. 
ReLU was selected as activation function $g$ of the hidden neurons of VAF sub-networks.  
Thus, for each dataset we obtained $4$ network models with $1$-hidden layer, and $6$ with $2$-hidden layers. 
Let us call $net_{m_1}$ and $net_{m_1,m_2}$ the $1$-hidden and $2$-hidden layer networks, respectively, with $m_1, m_2 \in \{10, 25, 50, 100\}$. 
On the basis of what was discussed in Section \ref{sec:systemArch}, to each network  $net_{m_1}$ ($net_{m_1,m_2}$)  it is possible to associate a neural network $vnet^k_{m_1}$ ($vnet^k_{m_1,m_2}$) equipped with  VAF subnetworks, where $k$ is the number of hidden neurons of VAF subnetworks. 

On the basis of the results discussed in Section \ref{subsec:preliminaryExp}, we considered VAF subnets shared on each layer, and $k=3$. 
In Table \ref{table:nnArch} we report the neural network architectures used in this series of experiments. 
Neural network architectures were sorted in ascending order according to their complexity.
Networks were trained according to an usual learning approach, described in Algorithm \ref{algo:standard-learning}. 
In particular, we used a batch approach, RProp \citep{riedmiller1992}, with ``small'' datasets, i.e, when the number of examples was less than $5 \cdot 10^3$, otherwise we used a mini-batch approach, RMSProp \citep{tieleman2012}. 
Moreover, networks with VAF subnetworks were trained using both a random initialization and an specific weight initialization such that they approximate a ReLU function.
All the network models, i.e., $net_{m_1}$,$net_{m_1,m_2}$,$vnet^k_{m_1}$ and $vnet^k_{m_1,m_2}$ were compared in a $K$-fold cross validation schema (see Algorithm \ref{algo:kfcv}), with $K=10$. 

Note that Learning Rate (LR) in RMSProp spans in the range $[0.0001,0.1]$, considering $10$ equispaced values, while in RProp $\eta^{+}$ was selected equal to $1.01$, and $\eta^{-}$ equal to $0.5$.  
In Table \ref{table:firstExpScenario} are summarized the parameters of this series of empirical evaluations. 


\begin{algorithm}[H]
\SetAlgoLined

\KwIn{Dataset $D$, network model $mnet$, number of folds $k$, hyper-parameters values $\{p_1,p_2,\dots,p_n\}$ with $p_i=\{$ possible values for $i$-th parameter $\}$ with $1\leq i\leq n$}

		$FoldResults=[ \ ]$\;
		split $D$ in a $k-$partition $P^{k}(D)$ \;

		\ForAll{$1 \leq i \leq k $}{								
				 $TestSet \leftarrow  P^{k}_i(D) $\;
				$R \leftarrow P^{k}(D)\setminus \{TestSet\}$\;
		
				split $R$ in a $2-$partition $P^{2}(R)$ \;
				$TrainSet \leftarrow  P^{2}_1(R)$\;
				$ValSet \leftarrow  P^{2}_2(R)$\;
				$bestParams \leftarrow \emptyset$\;
				$bestResults \leftarrow \emptyset$\;
				\ForAll{$h \in p_1 \times p_2\times \dots \times p_n $}	{
														
						$model\leftarrow Train(mnet,TrainSet,ValSet,h)$\;
						$results\leftarrow Sim(model, TestSet)$\;

					\If{$results$ \textbf{better than} $bestResults$}
                    {
						$bestResults \leftarrow results$\;
						$bestParams \leftarrow h$\;
						 $bestModel \leftarrow model$\;
					}
					
				}		
				$FoldResults[i]\leftarrow bestResults;$
		}					
		\textbf{return} $Average(FoldResults)$

 
 \caption{$K$-fold cross validation procedure}
 \label{algo:kfcv}
\end{algorithm}


\begin{table}[t]
\centering
\scalebox{0.7}
{
\begin{tabular}{l c c c c c l}

\hline
      Name & Istances & Input Dim. & N. classes & Task & Neural Network Arch. & Ref.\\
      \hline
      CPU-Small      &  8192 & 12 & - & Regress. & MLFF & \cite{dua2017} \\
      DeltaElevator  &  9517 & 6  & - & Regress. & MLFF & \cite{dcc}\\
      Elevators      & 16599 & 18 & - & Regress. & MLFF & \cite{dcc}\\
      Kinematics     &  8192 & 8  & - & Regress. & MLFF & \cite{dcc}\\
      Puma-8NH       &  8192 & 8  & - & Regress. & MLFF & \cite{dcc}\\
      Puma-32NH      &  8192 & 32 & - & Regress. & MLFF & \cite{dcc}\\
      Servo          &   197 &  4 & - & Regress. & MLFF & \cite{dua2017}\\
      Energy Cooling &   768 &  8 & - & Regress. & MLFF & \cite{dcc}\\
      Energy Heating &   768 &  8 & - & Regress. & MLFF & \cite{dcc}\\
      Yatch          &   308 &  7 & - & Regress. & MLFF & \cite{dua2017}\\
      \hline
      Sensorless    & $58509$ & $49$ & 11 & Classif & MLFF & \cite{dua2017}\\
      Liver & $345$ & $7$ & $2$ & Classif. & MLFF & \cite{dua2017}\\
      Wine & $178$ & $13$ & $3$ & Classif. & MLFF & \cite{dua2017}\\
      Statlog Image Segmentation & $2310$ & $19$ & $7$ & Classif. & MLFF & \cite{dua2017}\\
      Statlog Landsat Satellite  & $6435$ & $36$ & $7$ & Classif. & MLFF & \cite{dua2017}\\
      Cardiotocography           & $2126$ & $22$ & $3$ & Classif. & MLFF & \cite{dua2017}\\
  Seismic bumps & $2584$ & $18$ & $2$ & Classif. & MLFF & \cite{sikora2010}\\
Dermatology                &  $336$ & $35$ & $3$ & Classif. & MLFF & \cite{dua2017}\\
Diabetic retinopathy debrecen & $1151$ & $19$ & $2$ & Classif. & MLFF & \cite{antal2014}\\
QSAR biodegradation & $1055$ & $41$ & $2$ & Classif. & MLFF & \cite{mansouri2013}\\
Climate model simulation & $540$ & $18$ & $2$ & Classif. & MLFF & \cite{lucas2013}\\

  \hline
      MNIST                      & $70000$ & $28 \times 28$ & $10$ & Classif. & CNN & \cite{lecun2010} \\
      Fashion MNIST              & $70000$ & $28 \times 28$ & $10$ & Classif. & CNN & \cite{xiao2017} \\    
      Cifar10           & $60000$ & $32 \times 32 \times 3$ & $10$ & Classif. & CNN & \cite{krizhevsky2009}\\
  \hline

\end{tabular}
}
\caption{Properties of the datasets used for the experiments, and architectures of the neural network  applied to the data.}
\label{table:dataset}
\end{table}

\begin{table}[th!]
\centering
\small
\caption{Neural network architectures used in the first experimental scenario. See text for further details.}
\label{table:nnArch}
\begin{tabular}{@{}l|llllllllll@{}}
\toprule
                                                                                        &\#1 & \#2 & \#3 & \#4 & \#5 & \#6 & \#7 & \#8 & \#9  & \#10     \\ \midrule
Stand &$net_{10}$ & $net_{25}$& $net_{50}$& $net_{100}$& $net_{25,10}$& $net_{50,10}$& $net_{100,10}$& $net_{50,25}$& $net_{100,25}$&$net_{100,50}$ \\
VAF &$vnet^3_{10}$ & $vnet^3_{25}$& $vnet^3_{50}$& $vnet^3_{100}$& $vnet^3_{25,10}$& $vnet^3_{50,10}$& $vnet^3_{100,10}$& $vnet^3_{50,25}$& $vnet^3_{100,25}$&$vnet^3_{100,50}$ \\
\bottomrule
\end{tabular}
\end{table}

\begin{table}[th!]
\centering
\caption{Parameters of the first experimental scenario. See text for further details.}
\label{table:firstExpScenario}
\begin{tabular}{@{}llllll@{}}
\toprule
                                                                                        $m_1$, $m_2$ &  k & VAF initialization & Learning approaches & \# maximum epochs & K     \\ \midrule
\{10, 25, 50, 100\} & \{3\} & \{Random, ReLU\} & \{RMSProp, RProp\} & 300 & 10\\
\bottomrule
\end{tabular}
\end{table}

\subsubsection*{Results}
In Table \ref{tab:ris_reg} and \ref{tab:ris_clas} are showed mean and standard deviations of RMSE and accuracy for regression and classification datasets, respectively, by using a K-fold cross-validation approach. 
The best results are displayed in bold. 

In case of the regression datasets, VAF approach uniformly overcomes standard approach. 
Only in one case we obtain the best result with a standard approach. 
For four datasets (DeltaElevator, Elevators, Puma-32H and Yatch) RMSE 's mean obtained by VAF networks results much smaller than RMSE obtained by neural network without VAF subnetwork. 
For example in DeltaElevator dataset RMSE's mean was reduced by two (VAF init random) and one (VAF Init ReLU) order of magnitude. 
Moreover standard deviations remain comparable or lower than those without VAF subnetworks. 
This suggests that the training process of network with VAF subnetworks is sufficiently stable. 


Similar results were obtained with classification datasets (see Table \ref{tab:ris_clas}). 
Neural networks with VAF outperforms neural networks without VAF. 
Only in two datasets ($20 \%$) neural networks without VAF outperforms neural networks with VAF. Also in this case,  standard deviations remain comparable or lower than those without VAF.

\begin{figure}
    \centering
    \begin{subfigure}[b]{0.48\textwidth}
        \includegraphics[width=\textwidth]{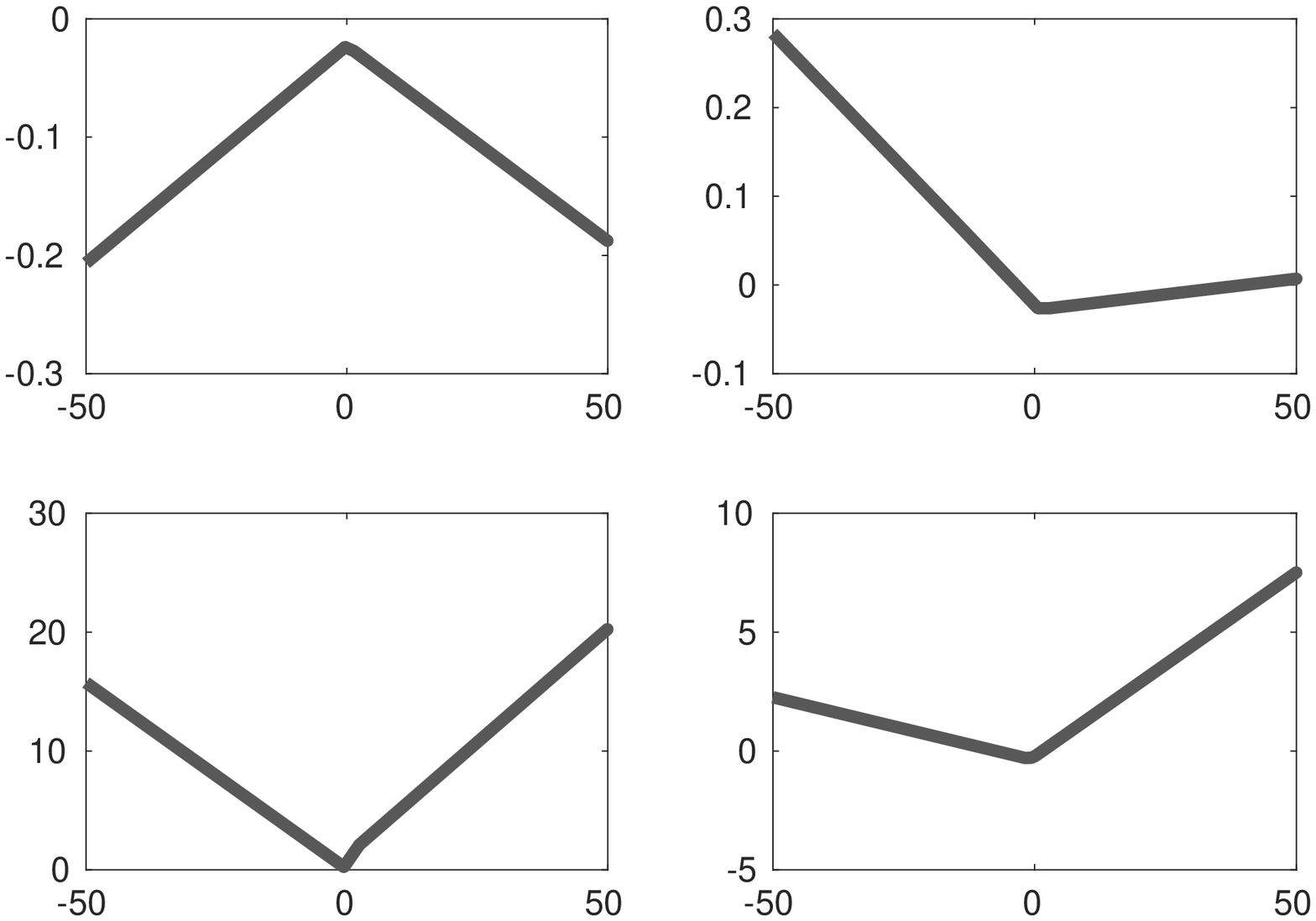}
        \caption{Regression}
        \label{subfig:regExampleVaf}
    \end{subfigure}
    \begin{subfigure}[b]{0.48\textwidth}
        \includegraphics[width=\textwidth]{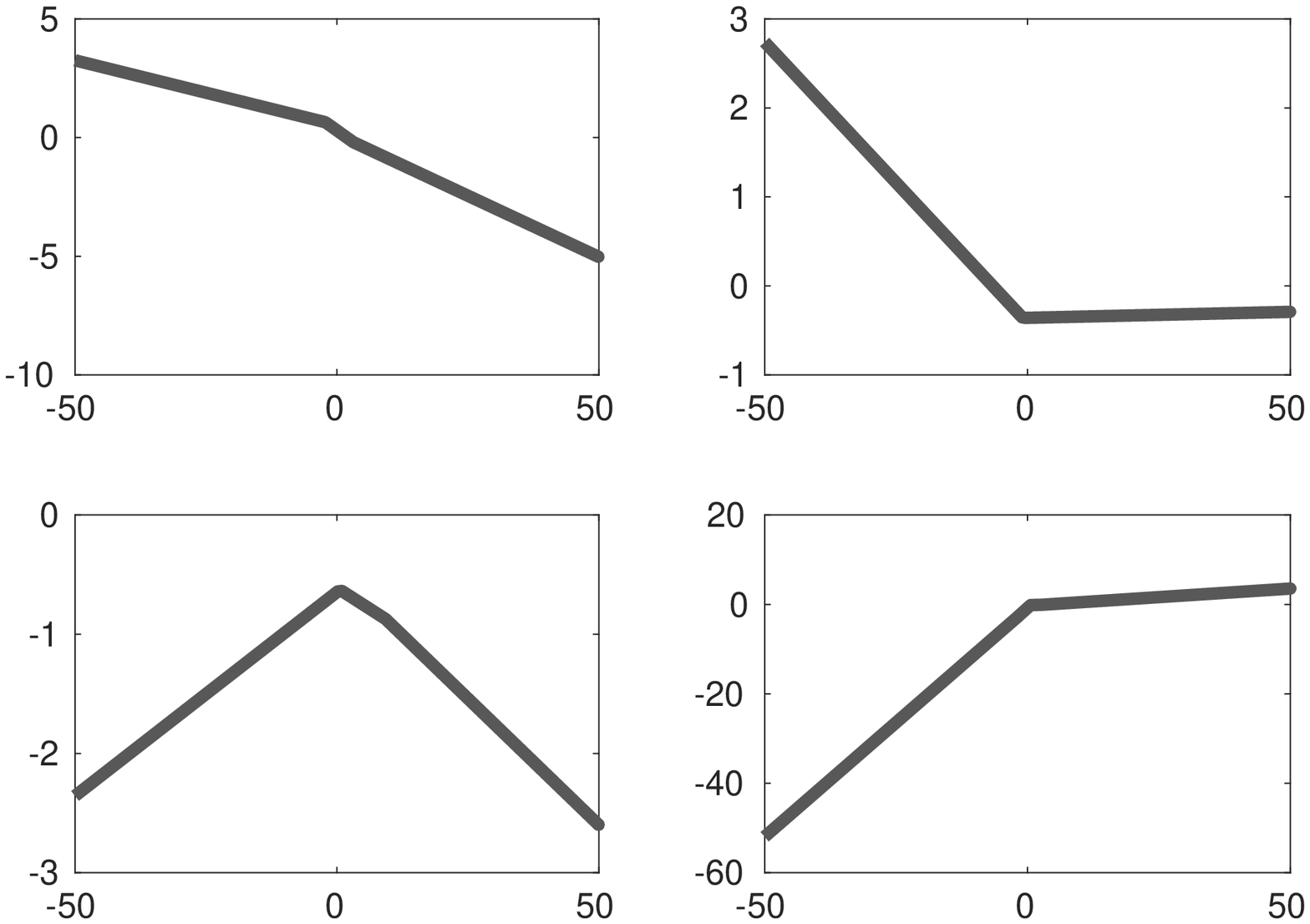}
        \caption{Classification}
        \label{subfig:claExampleVaf}
    \end{subfigure}
    \caption{Plots of some VAF behaviours at the end of the learning process. In \ref{subfig:regExampleVaf} for regression datasets, in \ref{subfig:claExampleVaf} for classification datasets.} 
    \label{fig:vafExamples2}
\end{figure}

In Figure \ref{fig:vafExamples} and \ref{fig:vafExamples2} are reported some VAF subnetwork behaviours at the end of the learning process.

\begin{table}[th!]
\centering
\resizebox{\textwidth}{!}{
\begin{tabular}{ccccc}
\hline
               & standard  Relu       & VAF Init random  & VAF Init ReLU    &  \\
               & RMSE: mean + St.Dev    & RMSE: mean + St.Dev    & RMSE: mean + St.Dev    &  \\
               \hline
CPUsmall       & $0.0616  \pm  0.0016$ ($net_{50,10}$)& $\mathbf{0.0593\pm 0.0017}$ ($vnet^3_{100}$)& $0.0606 \pm 0.0032$ ($vnet^3_{100,10}$)&  \\

DeltaElevator  & $0.1355  \pm  0.0055$ ($net_{25}$)& $\mathbf{0.0030\pm 0.0006}$ ($vnet^3_{100, 50}$)& $0.0414 \pm 0.0393$ ($vnet^3_{10}$)&  \\

Elevators      & $1.2746 \pm 0.3741$ ($net_{50,10}$)& $\mathbf{0.0068\pm 0.0003}$ ($vnet^3_{50,25}$)& $0.1915 \pm 0.1658$ ($vnet^3_{10}$)&  \\

Kinematics     & $0.1090 \pm 0.0040$ ($net_{100}$)& $0.1315\pm 0.0185$ ($vnet^3_{25}$)& $\mathbf{0.0935     \pm 0.0048}$ ($vnet^3_{50,25}$)&  \\

Puma-8NH       & $0.1336 \pm 0.0023$ ($net_{25, 10} $)& $\mathbf{0.1316\pm 0.0018}$ ($vnet^3_{100,10}$)& $0.1331 \pm 0.0025$ ($vnet^3_{25,10}$)&  \\

Puma-32H       & $0.2372 \pm 0.1473$ ($net_{25,10}$)& $\mathbf{0.0273\pm 0.0005}$ ($vnet^3_{100,10}$)& $0.0317 \pm 0.0039$ ($vnet^3_{25,10}$)&  \\

Servo          & $0.0946 \pm 0.0158$ ($net_{100}$)& $\mathbf{0.0896\pm 0.0276}$ ($vnet^3_{10}$)& $0.0961 \pm 0.0271$ ($vnet^3_{25}$)&  \\

Energy Cooling & $0.0417 \pm 0.0014$ ($net_{100,10}$)& $0.0461\pm 0.0024$ ($vnet^3_{100,25}$)& $\mathbf{0.0400     \pm 0.0026}$ ($vnet^3_{50,25}$)&  \\

Energy Heating & $\mathbf{0.0206 \pm 0.0026}$ ($net_{100,10}$)& $0.0304\pm 0.0237$ ($vnet^3_{25,10}$)& $0.0213 \pm 0.0027$ ($vnet^3_{100,10}$)&  \\

Yatch          & $0.2442 \pm 0.1146$ ($net_{25}$)& $0.3435 \pm 0.2186$ ($vnet^3_{100}$)& $\mathbf{0.1481 \pm 0.0553}$ ($vnet^3_{25}$)& \\
\hline
\end{tabular}
}
\caption{RMSE for the experiments on the regression datasets. We used a K-Fold Cross-validation evaluation. In bold the best results. The best neural architecture for each case is between parentheses.}
\label{tab:ris_reg}
\end{table}

\begin{table}[th!]
\centering
\resizebox{\textwidth}{!}{
\begin{tabular}{ccccc}
\hline
               & standard  Relu       & VAF Init random  & VAF Init ReLU    &  \\
               & Accuracy: mean + St.Dev    & Accuracy: mean + St.Dev    & Accuracy: mean + St.Dev    &  \\
               \hline
Liver       & $0.6203  \pm 0.0474$ ($net_{25,10}$)& $0.6290  \pm 0.0378$ ($vnet^3_ {100,10}$)& $\mathbf{0.6348  \pm 0.0375}$($vnet^3_{25}$)&  \\

Wine  & $0.8879 \pm 0.0516$ ($net_{10} $)& $\mathbf{0.9552 \pm 0.0371}$ ($vnet^3_{50} $)& $0.9162 \pm 0.0434$ ($vnet^3_{10}$)&  \\

Image segmentation      & $\mathbf{0.9463 \pm 0.0128}$ ($net_{25} $)& $0.9351 \pm 0.0179$ ($vnet^3_{50}$)& $0.9381  \pm 0.0079$ ($vnet^3_{50}$)&  \\

Satellite image     & $0.8821 \pm 0.0101$ ($net_{100,50} $)& $0.8856 \pm 0.0028$ ($vnet^3_{100} $)& $\mathbf{0.8875 \pm 0.0080}$ ($vnet^3_{100,25}$)&  \\

CTG       & $0.8979  \pm  0.0263$ ($net_{100} $)& $\mathbf{0.9040  \pm  0.0073}$ ($vnet^3_{100} $)& $0.8984  \pm  0.0261$ ($vnet^3_{50,25}$)&  \\

Seismic bumps       & $\mathbf{0.9346 \pm 0.0009}$ ($net_{10} $)& $0.9234 \pm 0.0074$ ($vnet^3_{10} $)& $0.9342 \pm 0.0001$ ($vnet^3_{10}$)&  \\

Dermatology          & $0.9749 \pm  0.0116$ ($net_{50,25} $)& $0.9692 \pm  0.0182$ ($vnet^3_{10} $)& $\mathbf{0.9750  \pm 0.0248}$ ($vnet^3_{100}$)&  \\

Diabetic & $0.7254 \pm  0.0290$ ($net_{100} $)& $0.7315 \pm  0.0238$ ($vnet^3_{10} $)& $\mathbf{0.7333  \pm 0.0231}$ ($vnet^3_{50}$)&  \\

Biodegradation & $0.8635 \pm 0.0336$ ($net_{10} $)& $\mathbf{0.8673 \pm 0.0225}$ ($vnet^3_{100,10} $)& $0.8569 \pm 0.0108$ ($vnet^3_{50}$)&  \\

Climate simulation          & $0.9500 \pm 0.0140$ ($net_{50,25} $)& $0.9519 \pm 0.0211$ ($vnet^3_{10} $)& $\mathbf{0.9556 \pm  0.0240}$ ($vnet^3_{100}$)& \\
	    	    
\hline
\end{tabular}
}
\caption{Accuracies for the experiments on the classification datasets. We used a K-Fold Cross-validation evaluation. In bold the best results. The best neural architecture for each case is between parentheses.}
\label{tab:ris_clas}
\end{table}

\subsection{Convolutional MLFF networks}
\label{subsubsection:secondExp}
In order to evaluate experimentally the impact of VAF on Convolutional Neural Networks (CNN), we consider standard CNN networks with $2$ and $3$ convolutional layers, and run experiments on three different dataset:  MNIST, Fashion MNIST and CIFAR10 (see Table \ref{table:dataset} for further details). 
As discussed in Section \ref{subsec:vafLearning} and Section \ref{subsubsection:firstExp}, a key aspect is the initialization of the VAF networks. 
Thus, also in this case, we chose to initialize the weights of the VAF subnetworks either randomly or to approximate a ReLU function. To this aim, we build two CNN architectures (similar to the basic network used in \citep{lin2013}), the first one composed of $2$-layer CNN networks used for MNIST and Fashion-MNIST and the second one composed of $3$-layers trained and tested with the more complex CIFAR10 dataset.
Let us call $cnet_{A_2}$  and  $cnet_{A_3}$ respectively the $2$-layer CNN and the $3$-layer CNN; as stated in Section \ref{sec:systemArch}, it is possible to associate to each $cnet$ a neural network $vcnet^k$ equipped with VAF sub-networks having $k$ hidden units. 
The experiments were preformed using a  $10$-fold cross validation schema as described in \ref{algo:kfcv}.
Networks were trained using Stochastic Gradient Descent (SGD) method with mini-batching. 

Furthermore, we compare our architecture with two 
other neural architectures also equipped with trainable activation functions. The first one is KAFnet, a very recent and promising approach proposed in  \citep{scardapane2018} and  already discussed in Section \ref{subsec:linearCombinationBF}. The second one is Network in Network (NIN), a successful approach proposed in \citep{lin2013} and already discussed in Section \ref{subsec:otherApproaches}.
To this aim, we used the same experimental settings described  in \citep{scardapane2018}, i.e., a convolutional MLFF network composed by two convolutional layers, each of these followed by a $3\times 3$ maxpooling layer and a dropout layer of $0.25$ (see Table \ref{table:conv_arch}). To distinguish it from the others models, we will call this network $cnet_B$.
Starting from $cnet_B$, we obtained three different types of neural networks with trainable activation functions according to three different procedures proposed for KAFnet, NIN and VAF.
We use the classic CIFAR10 data configuration ($50000$ training samples + $10000$ test samples) to train the three types of obtained networks.
The network $cnet_B$ with fixed activation function corresponding to ReLU is also considered as baseline.
Finally, we repeat the same setup using MNIST and Fashion-MNIST dataset.

Properties of the used CNN architectures and learning process are summarised in table \ref{table:conv_arch}.

\begin{table}[th!]
\centering
\caption{Parameters of the second experimental scenario. See text for further details.}
\scalebox{0.8}{
\begin{tabular}{@{}lllllll@{}}
\toprule
Name & Layers & VAF initialization & Learning approaches & \# maximum epochs    \\ \midrule
$cnet_{A_2}$ & $2\times$ (Conv. $192$  + Maxout + Dropout)  & \{Random, ReLU\} & SGD & 300 \\
$cnet_{A_3}$ & $3\times$ (Conv. $192$ + Maxout + Dropout) & \{Random, ReLU\} & SGD & 300 \\
$cnet_B$ & $2 \times$ (Conv. $150$ + Maxout + Dropout)  & Random & Adam & 300\\

\bottomrule

\end{tabular}
}
\label{table:conv_arch}
\end{table}


\subsubsection*{Results}
\label{subsubsec:results}
In Table \ref{tab:ris_exp2a} are shown mean and standard deviations of accuracies for the three datasets  Cifar10, MNIST and Fashion MNIST, using a $10$-fold cross-validation approach for the neural architecture summarized in the first two rows of Table \ref{table:conv_arch}. 
The best results are reported in bold style. One can note that VAF approach uniformly outperforms the standard approach, especially when using a random initialization scheme. 
Also in this experimental scenario the standard deviations obtained by networks with VAF remain comparable or lower than those without VAF subnetworks. 
Especially for the CIFAR10 dataset, we obtain a considerable improvement.\\

\begin{figure}[t]
\centering
  \begin{minipage}[h]{.7\textwidth}
  \includegraphics[width=0.46\textwidth]{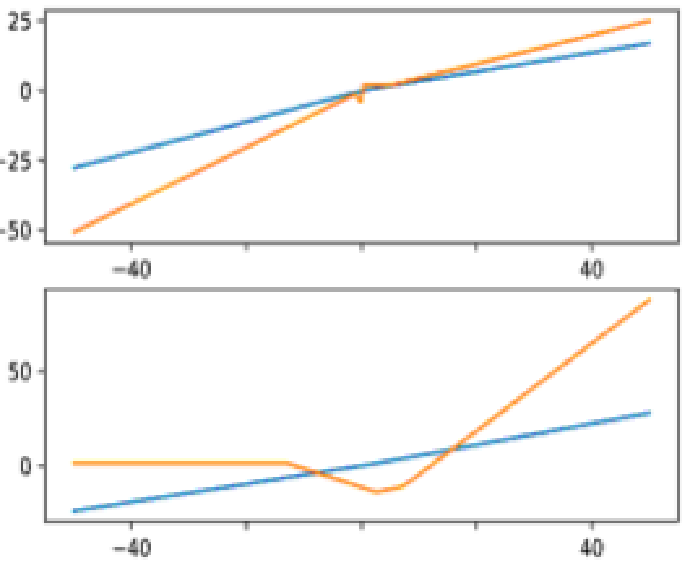}
  \includegraphics[width=0.48\textwidth]{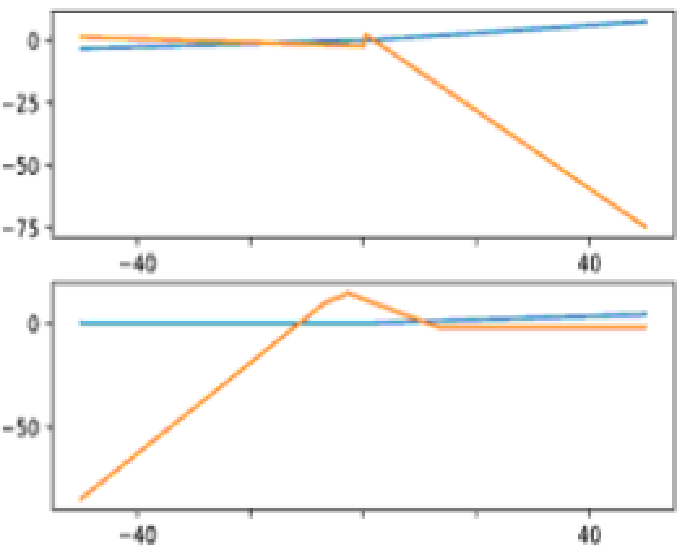}

  \includegraphics[width=0.46\textwidth]{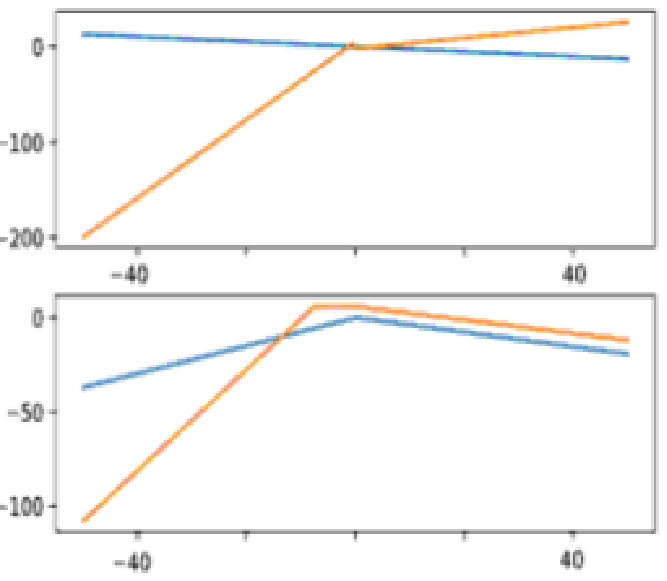}
  \includegraphics[width=0.48\textwidth]{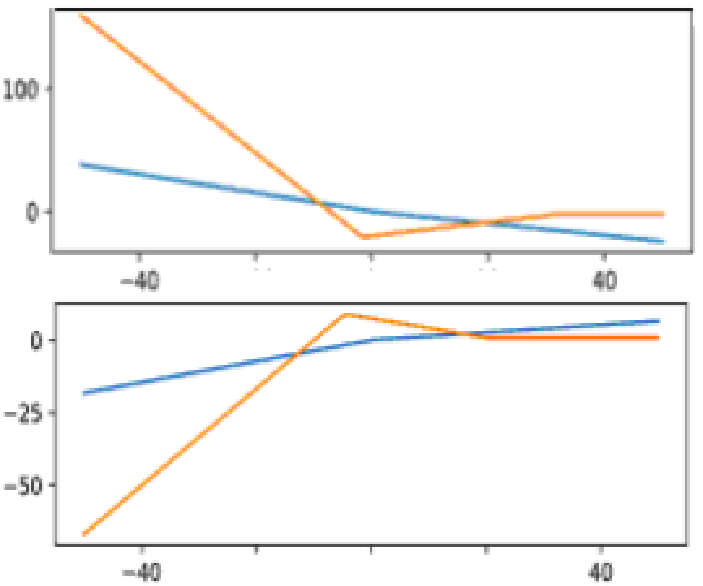}
  \caption*{Random init.}
  \end{minipage}
  \begin{minipage}[h]{.7\textwidth}
  \includegraphics[width=0.48\textwidth]{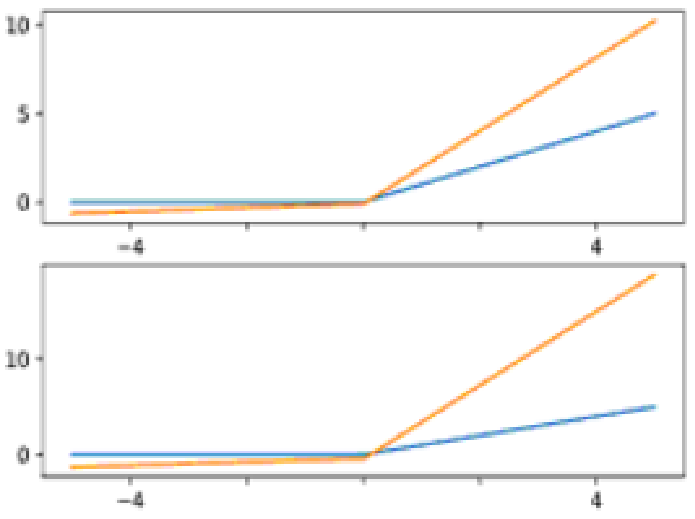}
  \includegraphics[width=0.48\textwidth]{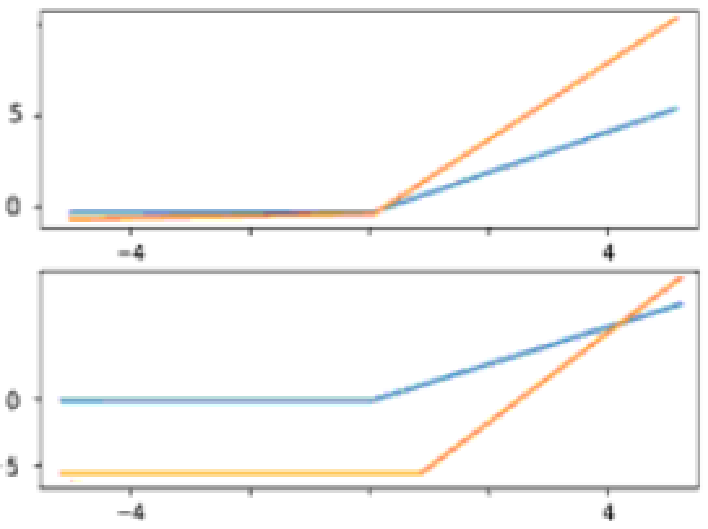}

  \includegraphics[width=0.48\textwidth]{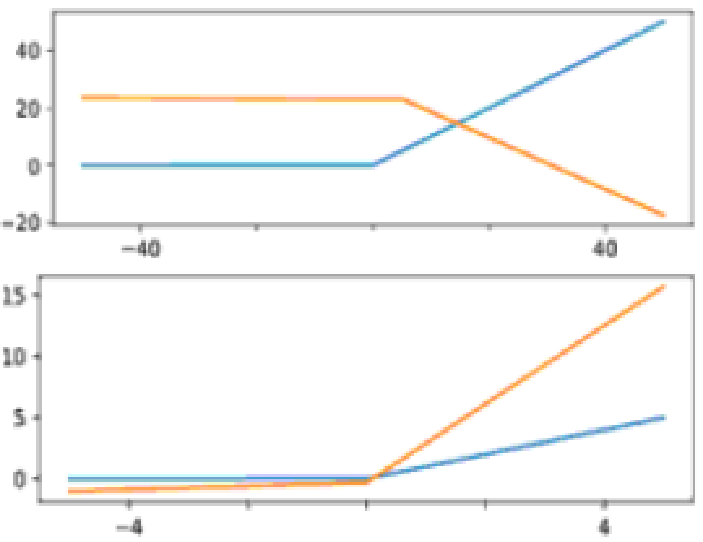}
  \includegraphics[width=0.48\textwidth]{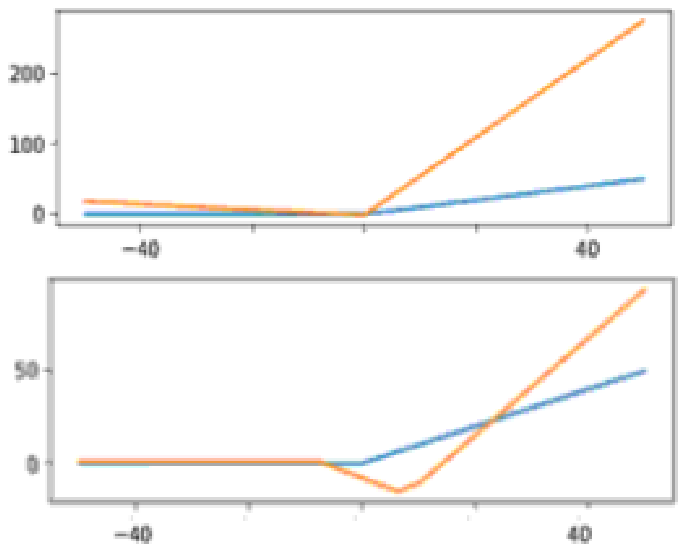}
  \caption*{ReLU init.}

  \end{minipage}
\caption{Examples of changes in a VAF in a 2 layer conv. network using random (top) and ReLU initialization (bottom). The blue line is the start function, the orange line is the learned function. }
\label{fig:lfr2s}
\end{figure}

\begin{figure}[t]
	\centering
  \begin{minipage}[h]{0.9\textwidth}
  \includegraphics[width=0.48\textwidth]{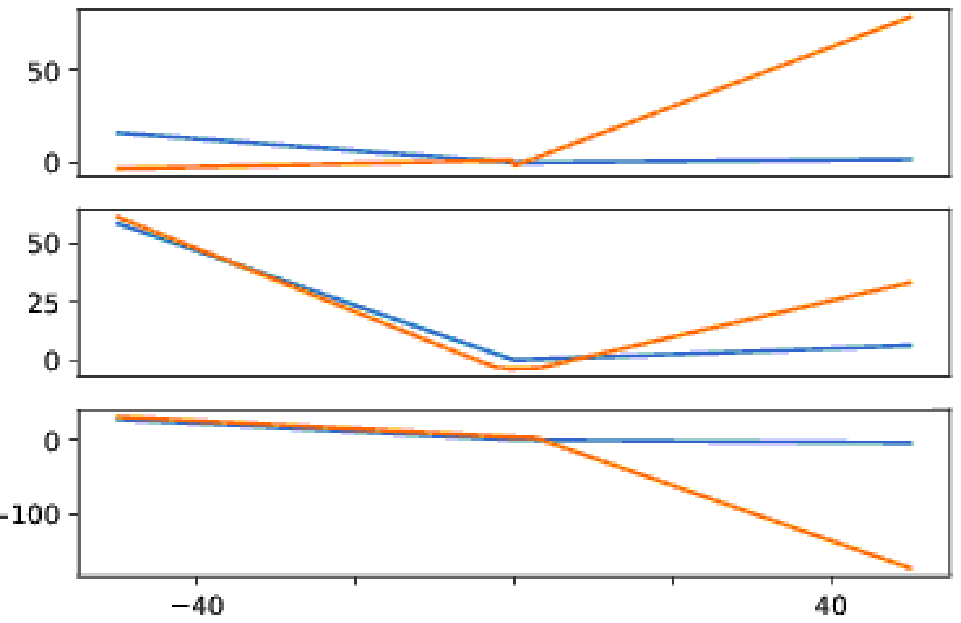}
  \includegraphics[width=0.48\textwidth]{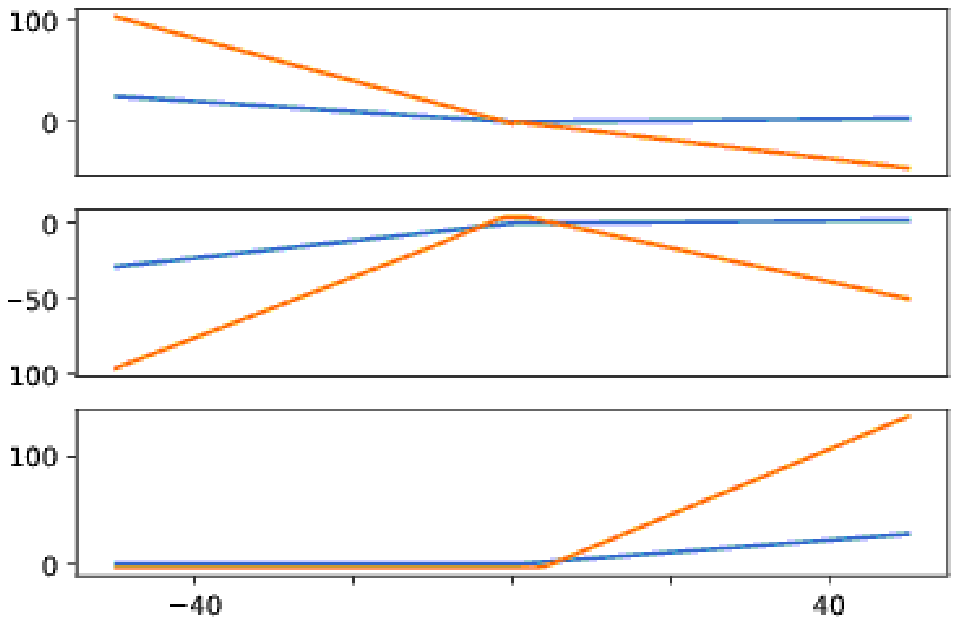}
  \caption*{Random init.}
  \end{minipage}
  \begin{minipage}[h]{0.9\textwidth}
  \includegraphics[width=0.48\textwidth]{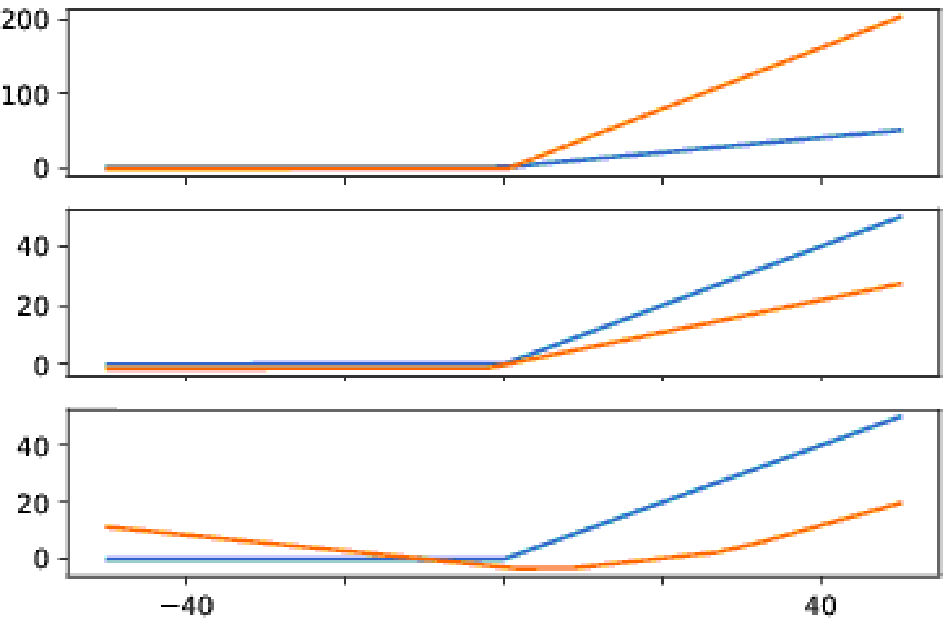}
  \includegraphics[width=0.48\textwidth]{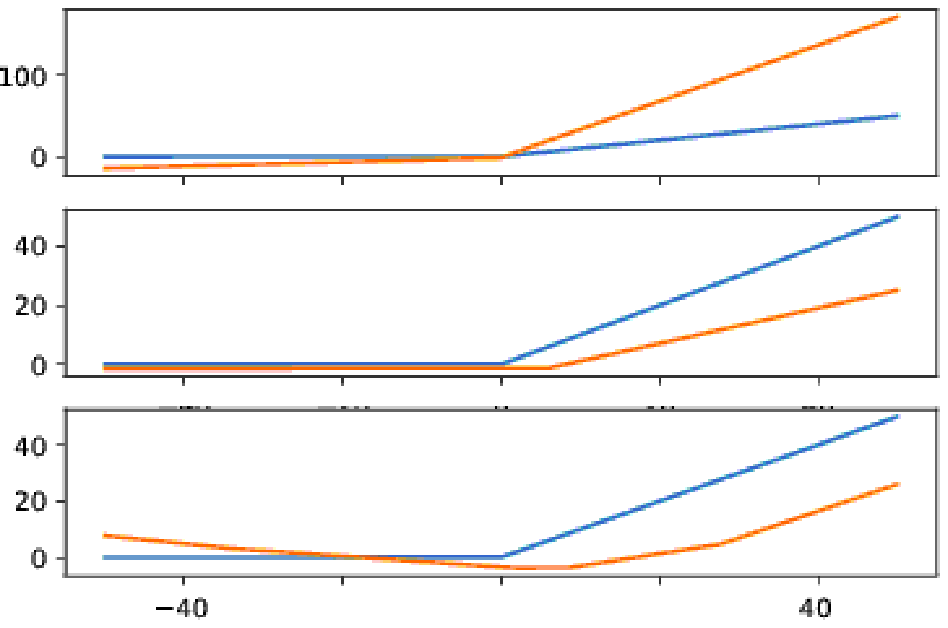}
  \caption*{ReLU init.}
  \end{minipage}
\caption{Examples of a resulting VAF in a 3 layer conv. using random (top) and ReLU initialization (bottom). The blue line is the start function, the orange line is the learned function.}
\label{fig:lfr3s}
\end{figure}

In Figures \ref{fig:lfr2s} and \ref{fig:lfr3s} are shown some examples of trained activation functions respectively in $vcnn_{A_2}$ and $vcnn_{A_3}$; it should be noted the influence of VAF initialization on the trained activation function: it seems that, in case of initialization as ReLU, the initial shape remains mostly unchanged, giving a resulting function that looks like a PReLU/Leaky ReLU. A more interesting behaviour is given by random initialization, where every VAF unit seems to exhibit greater changes respect to the initial function. 
This greater variability given by random initialization respect to ReLU initialization seems to give an improvement in accuracy results as shown in Table \ref{tab:ris_exp2a}.

In Table \ref{tab:ris_exp2b} we show the performances of KAFnet, NIN and VAF network on the two datasets Fashion-MNIST and CIFAR10  \ref{table:conv_arch} in terms of accuracy. VAF network outperforms KAF and NIN on both the dataset CIFAR10 and Fashion MNIST. We do not report the MNIST results because are all very similar between them (over the $99\%$ of accuracy). Notably, therefore, also with respect to two other two approaches with trainable activation functions known in literature, our approach results in better performance.
\begin{table}[th!]
\centering
\begin{tabular}{ccccc}
\hline
& standard  ReLU       & VAF Init random  & VAF Init ReLU    &  \\
& Acc. + St.Dev    & Acc. + St.Dev    & Acc. + St.Dev    &  \\
\hline
Cifar10  & $0.857 \pm 0.002$ $(cnet^5_3)$ &  $\mathbf{0.875 \pm 0.003}$  $(vcnet^5_{A_3})$ & $0.860 \pm 0.002$  $(vcnet^5_{A_3})$ & \\
MNIST  & $0.991\pm 0.001$  $(cnet^5_{A_2})$ & $\mathbf{0.994\pm 0.001}$ $(vcnet^5_{A_2})$ & $0.993\pm 0.002$ $(vcnet^5_{A_2})$ &  \\
Fashion MNIST & $0.923\pm 0.001$ $(cnet^5_{A_2})$ & $\mathbf{0.935\pm 0.002}$ $(vcnet^5_{A_2})$ & $0.934 \pm 0.001$ $(vcnet^5_{A_2})$ &  \\
\hline
\end{tabular}
\caption{Results of the convolutional networks with a $10$-fold cross Validation with $cnet_{A}$.}
\label{tab:ris_exp2a}
\end{table}

\begin{table}[th!]
\centering
\scalebox{1.0}{
\begin{tabular}{ccccc}
\hline
& standard ReLU & VAF(M=5) & KAF(D=20) & NIN  \\
& Accuracy & Accuracy & Accuracy  & Accuracy \\
\hline
Cifar10  & $0.707$ & $\mathbf{0.812}$ & $0.802$ & $0.763$\\
MNIST    & $0.995$ & $0.995$ &  $0.995$  &  $\mathbf{0.996}$ \\ 
Fashion MNIST & $0.920$ & $\mathbf{0.935}$ & $0.929$ &  $0.925$\\
\hline
\end{tabular}
}
\caption{Comparison between different activation functions on different dataset using the standard division on $cnet_B$.}
\label{tab:ris_exp2b}
\end{table}

\section{Conclusion}
\label{sec:conclusion}
In this work, we proposed a simple and direct way to obtain adaptable activation functions in feed-forward neural networks. 
In particular, we proposed to modify a feed-forward neural network by adding  Variable Activation Functions (VAF) in terms of one-hidden layer subnetworks (see Section \ref{sec:systemArch}). 
The resulting network is still a feed-forward neural network.  
The proposed architecture doesn't need many more parameters than networks using not adaptable activation functions as ReLU, and the learning process follows standard approaches (see Section \ref{subsec:vafLearning}). 
Importantly, VAF subnetworks can approximate arbitrarily well any activation functions, provided that the number of hidden neurons is sufficiently large (see Section \ref{sec:systemArch}).

It is worth to remark that our approach distinguishes from other approaches proposed in literature insofar as it satisfies simultaneously the properties \textit{p1} \,--\,\textit{p4} as described in Section \ref{subsec:summarizinApproaches}. These properties include a high expressive power of the trainable activation functions,  
no external parameter or learning process in addition to the classical ones for neural networks, and the possibility to use classical regularization methods. 

Interestingly, as we discussed in Section \ref{sec:systemArch} our architecture represents a general framework in which all the approaches described in Section \ref{subsec:linearCombinationBF} and some of the approaches in Section \ref{subsec:ensemble} can be included.

We experimentally evaluated our architecture on three different sets of experiments. In the former (see Section \ref{subsec:preliminaryExp}, we tested our approach using small shallow networks for defining some heuristic choices about VAF subnets. 
Notably, all the models equipped with VAF subnets outperform the corresponding shallow networks, and the results support the possibility of using a shared VAF approach with a fairly low number of VAF hidden neurons.
In the second series of experiments (see Section \ref{subsubsection:firstExp}), we considered full-connected Multi-Layered Neural Network (MLFF) networks. 
More specifically, we selected $10$ networks with $1$ or $2$ hidden layers. 
A correspondent network with VAF subnetworks was built for each of these $10$ networks (see Section \ref{sec:systemArch} and \ref{subsubsection:firstExp}). 
We obtained a total of $20$ different neural network architectures. 
These neural architectures were evaluated and compared using a $K$-Fold Cross-Validation procedure (see Algorithm \ref{algo:kfcv}) on $20$ different datasets (see Table \ref{table:dataset}), either for classification tasks or regression tasks. 
The results show that the networks with VAF subnetworks are uniformly more performing than the ones without VAF networks. 
In particular, our approach outperforms that without VAF networks on the $85 \%$ of the datasets. 
Only on three datasets our approach had worse results. 

In the last set of experiments, we considered Convolutional Neural Networks with $2$ and $3$ layers and correspondent networks with VAF units and we evaluate them using $3$ image datasets (MNIST, Fashion MNIST and CIFAR10)for classification. 
Also in this case the VAF subnetworks outperform networks with static units and selected state-of-the-art neural architectures (KAFNet and NIN) equipped with trainable activation functions.

In conclusion, VAF units have been tested using traditional MLNN networks and CNN networks with various datasets and give better results compared with networks with similar design both with traditional ReLU functions and trainable activation functions.
We showed that is possible to obtain  encouraging results without the need to use complex designs, particular initialization schemes or learning process in addition to those classically used for neural networks. 

\section*{Acknowledgements}
The work has been partially supported by the national project Perception, Performativity and Cognitive Sciences - PRIN2015 Cod. 2015TM24JS\_009.

\bibliographystyle{unsrtnat} 
\bibliography{bibliografia}
\newpage

\end{document}